\documentclass[11pt]{article}

\usepackage[final]{acl}

\usepackage{times}
\usepackage{latexsym}
\usepackage[table]{xcolor}
\definecolor{bbb}{RGB}{107, 154, 219}
\usepackage[T1]{fontenc}

\usepackage[utf8]{inputenc}
\definecolor{bluegradB}{RGB}{10, 78, 180}
\definecolor{bluegradL}{RGB}{25, 112, 210}
\definecolor{bluegradU}{RGB}{55, 150, 230}
\definecolor{bluegradE}{RGB}{95, 185, 245}
\newcommand{\blueword}{\textbf{\textcolor{bluegradB}{B}\textcolor{bluegradL}{L}\textcolor{bluegradU}{U}\textcolor{bluegradE}{E}}}
\usepackage{microtype}

\usepackage{inconsolata}

\usepackage{graphicx}
\usepackage{xcolor}
\usepackage[utf8]{inputenc}
\usepackage[T1]{fontenc}
\usepackage{hyperref}
\usepackage{url}
\usepackage{booktabs}
\usepackage{amsfonts}
\usepackage{amsmath}
\usepackage{amssymb}
\usepackage{bm}
\usepackage{nicefrac}
\usepackage{microtype}
\usepackage{xcolor}
\usepackage{graphicx}
\usepackage{algorithm}
\usepackage{algpseudocode}
\usepackage{multirow}
\usepackage{bbm}
\usepackage{mdframed}
\usepackage{enumitem}
\usepackage{wrapfig}
\usepackage{tikz}
\usepackage{etoc}
\usepackage{fontawesome5}
\newcommand{\corrauth}{\textsuperscript{\ensuremath{\dagger}}}

\newcommand{\cmark}{%
  \tikz[baseline=-0.01ex, x=0.24ex, y=0.24ex]{%
    \draw[line width=0.20ex, line cap=round, line join=round]
      (0,2) -- (2,0) -- (6,5);%
  }%
}
\newcommand{\xmark}{%
  \tikz[baseline=-0.01ex, x=0.24ex, y=0.24ex]{%
    \draw[line width=0.10ex, line cap=round] (0,0) -- (5,5);%
    \draw[line width=0.10ex, line cap=round] (0,5) -- (5,0);%
  }%
}

\title{\blueword: Toward \underline{\textcolor{bluegradB}{B}}etter \underline{\textcolor{bluegradL}{L}}anguage \underline{\textcolor{bluegradU}{U}}se in \underline{\textcolor{bluegradE}{E}}fficient \\Vision-Language-Action Models for Autonomous Driving}

\author{George Ling, Lijin Yang, Hao Yang, Zhongzhan Huang\corrauth \\
  Bosch Research \quad
\corrauth Correspondence to: \href{mailto:hru4sgh@bosch.com}{hru4sgh@bosch.com}\\
\href{https://github.com/George-Ling3/BLUE}{\faGithub\ Code} \quad 
\href{https://huggingface.co/George-Ling/blue_gate}{\faDatabase\ Hugging Face (Data/Logs/Ckpts)} \quad \href{https://blue-website.github.io}{\faGlobe\ Homepage}
}

\begin{document}
\etocdepthtag{main}
\maketitle
\begin{abstract}
We present \blueword, a minimal method for better language use in vision-language-action (VLA) models for autonomous driving (AD). 
Through extensive analysis, 
we reveal that language matters on only a small fraction of routes, but on those routes it can greatly improve or degrade performance. Generating language at every frame is therefore inefficient, since most computation is spent on frames that do not benefit from language.
We further show that pretrained VLA hidden states potentially already encode whether language will benefit a given frame, even though scene complexity and kinematic features alone struggle to predict this.
Based on this finding, BLUE trains a lightweight gate on frozen VLA hidden states to decide per frame whether to activate language generation or predict actions directly, without modifying the backbone or requiring additional human annotation. 
With just a 0.11M-parameter gate, BLUE sets a new state of the art on both benchmarks, achieving 76.2\% success rate on Bench2Drive and 36 driving score on Longest6 v2, while delivering 2.54$\times$ inference speedup and 8.9\% success rate improvement over the backbone.
BLUE provides a practical path toward efficient language-augmented AD, showing that VLA models can retain the benefits of language at a fraction of the cost.
\textbf{Our code, data, logs and checkpoints are fully available on \href{https://github.com/George-Ling3/BLUE}{Github}}.

\end{abstract}

\section{Introduction}

\begin{figure}[t]
  \vspace{-12pt}
  \centering
  \includegraphics[width=0.98\linewidth]{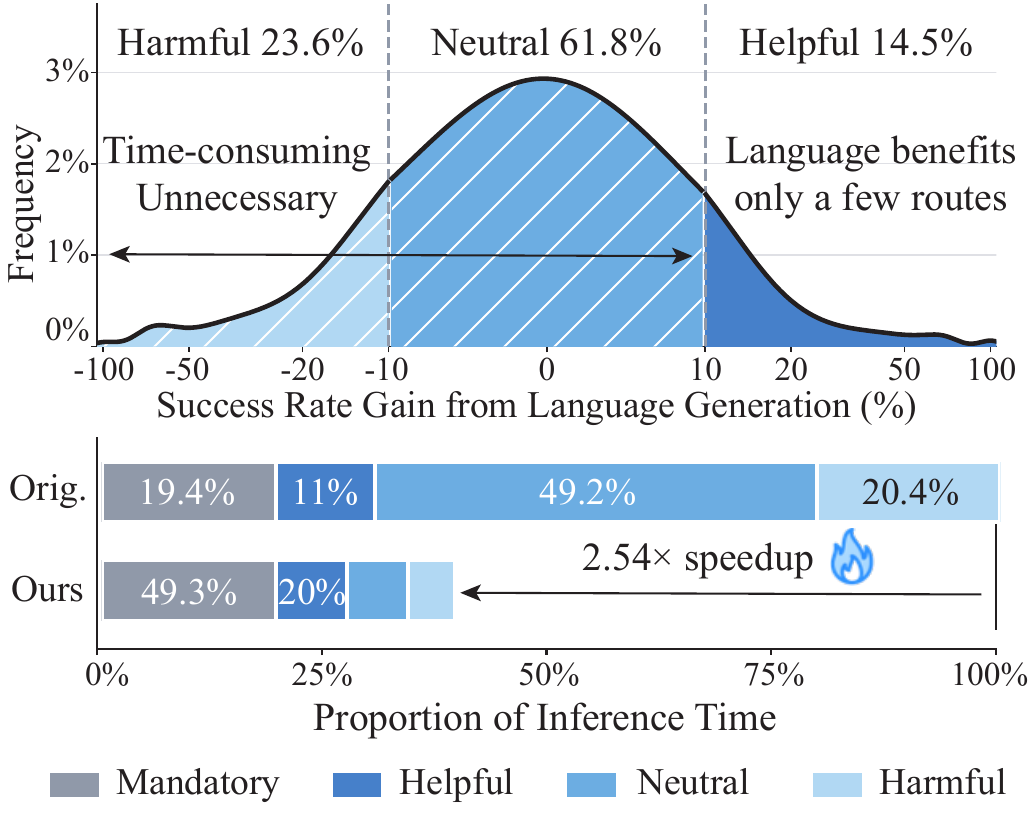}
  \vspace{-8pt}
  \caption{Top: Distribution of Bench2Drive routes by language impact. Bottom: Inference time breakdown comparing the original VLA and BLUE, which reduces unnecessary language generation for 2.54$\times$ speedup. 
  Extended details and results are in Appendix~\ref{sec:appendix_extra_settings}.
  }
  \label{fig:cot_analysis}
  \vspace{-12pt}
\end{figure}

Recent vision-language-action (VLA) models for autonomous driving typically generate natural language to reason about the scene before predicting driving actions \cite{renz2025simlingo, yang2026judge}. However, the impact of generated language on closed-loop driving has rarely been systematically quantified. 
First, we conduct $\sim$2000 GPU hours of closed-loop analysis on full Bench2Drive \cite{jia2024bench2drive} using SimLingo \cite{renz2025simlingo}, a representative VLA driving model, running each route through repeated experiments and categorizing driving outcomes via statistical tests.  As Figure~\ref{fig:cot_analysis} shows, the generated language statistically improves driving on only 14.5\% of routes, actively degrades it on 23.6\%, and has no clear effect on the remaining majority. Yet current many VLA driving models \cite{renz2025simlingo, yang2026judge, gao2026learning} usually generate language at every frame by default, wasting computation on frames that do not benefit and compromising both driving performance and inference efficiency. See more analysis on extra settings in Appendix~\ref{sec:appendix_extra_settings}.

\begin{figure*}[t]
  \centering
  \includegraphics[width=0.98\textwidth]{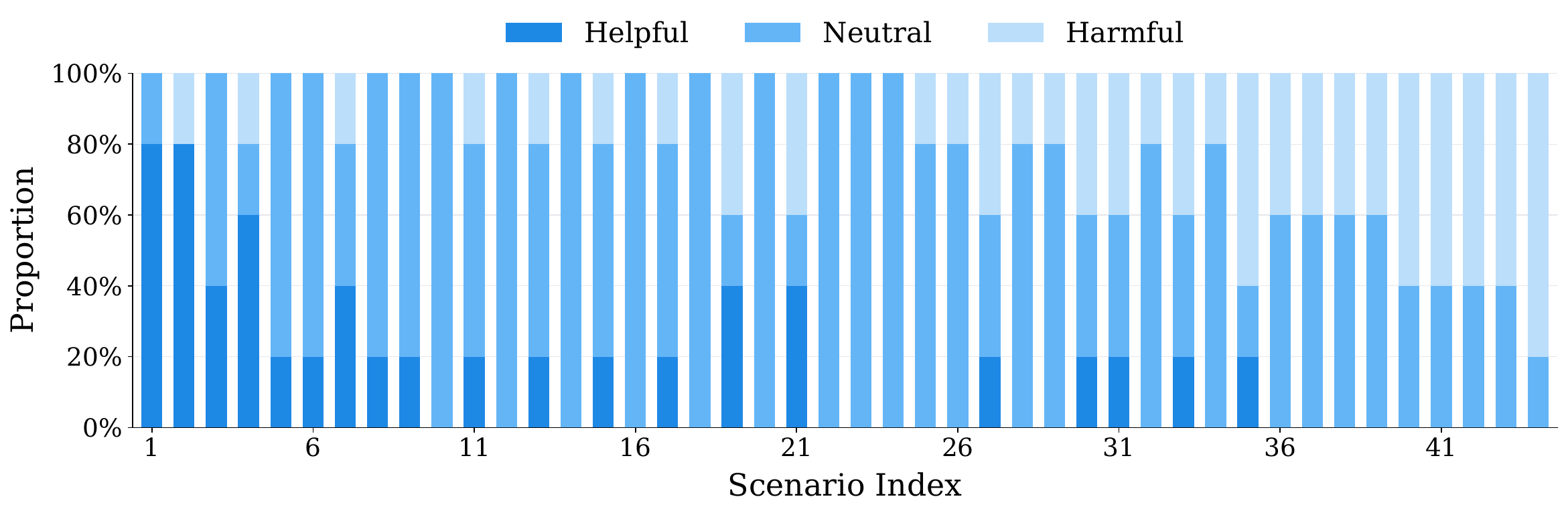}
  \vspace{-10pt}
  \caption{Per-scenario distribution of language effects across all 44 Bench2Drive scenario categories. Each bar represents one scenario, showing the proportion of routes where language generation is helpful, neutral, or harmful to driving success. Extended details and results under additional settings are in Appendix~\ref{sec:appendix_extra_settings}.}
  \label{fig:when_analysis}
  \vspace{-5pt}
\end{figure*}

Since language significantly helps or hurts driving performance on only a minority of routes at substantial inference cost, an intuitive strategy is to detect per frame whether language generation is needed and skip it otherwise, thereby improving both driving performance and inference speed.
Fortunately, we find that pretrained VLA hidden states potentially already encode this language-utility signal, even though scene complexity and kinematic features alone struggle to predict it. In Section~\ref{sec:method}, we leverage this finding and propose BLUE, a minimal method that trains a lightweight gate on frozen hidden states to decide per frame whether to activate language generation or predict actions directly. BLUE requires no backbone modification and no additional human annotation, as training labels are naturally derived from routine driving evaluation.
In Section~\ref{sec:experiments}, we show that with just a 0.11M-parameter gate, BLUE sets new state of the art on two benchmarks, achieving 76.2\% success rate on the multi-scenario Bench2Drive and 36 driving score on the long-horizon Longest6 v2, while delivering 2.54$\times$ inference speedup and a 8.9\% success rate improvement over the VLA backbone. We discuss related work in Appendix~\ref{sec:appendix_related_work} and summarize our contributions as follows:

\begin{itemize}[leftmargin=15pt]
\vspace{-5pt}
    \item We provide a systematic analysis of when language helps and when it hurts driving, showing that on-demand language use can improve both driving performance and inference speed.
    \item We reveal that pretrained VLA hidden states potentially already encode the language-utility signal. With just a 0.11M-parameter gate on frozen hidden states, BLUE achieves state-of-the-art performance on Bench2Drive and Longest6 v2, while delivering 2.54$\times$ inference speedup.
\end{itemize}

\section{Observations and Motivations}\label{sec:when}
We analyze how generated language affects driving performance across different scenarios and quantify the potential gains from selective language use.

\begin{figure*}[t]
  \centering
  \includegraphics[width=0.95\textwidth]{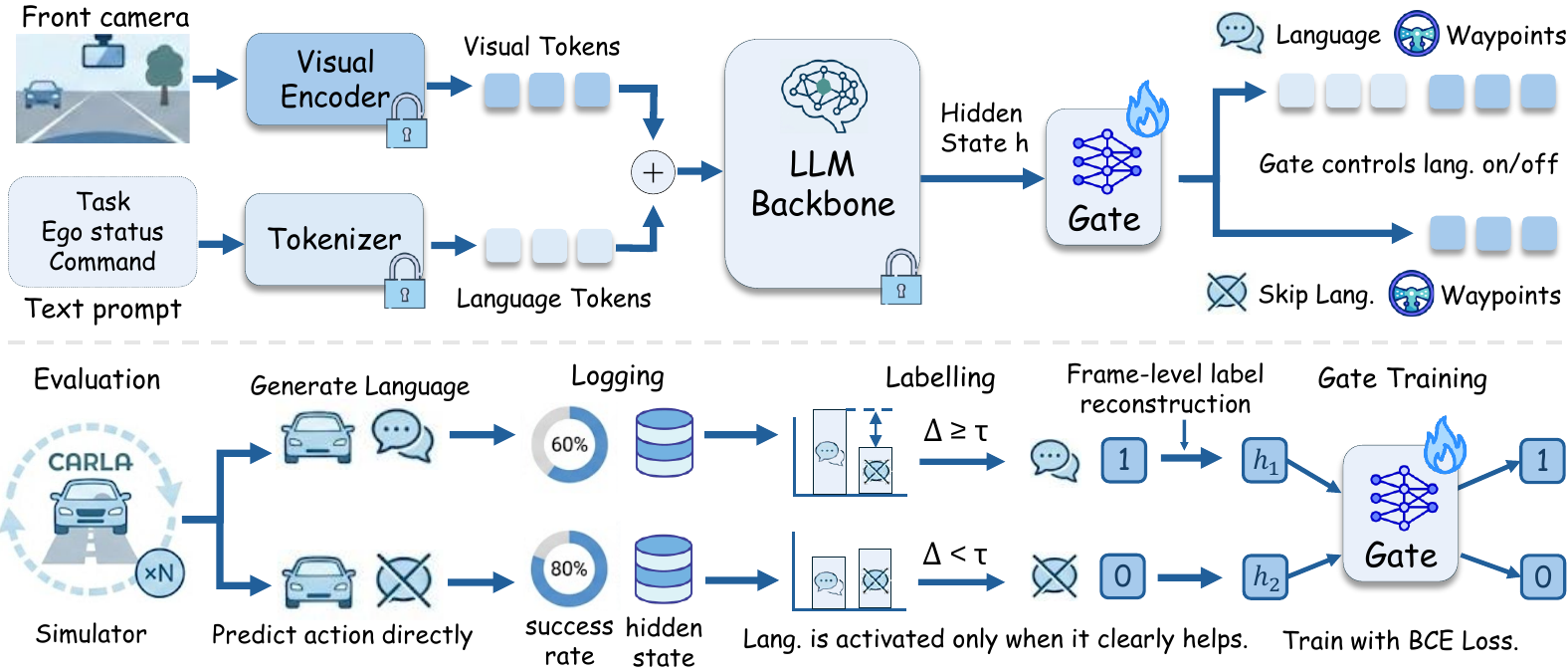}
  \caption{Overview of BLUE. \textbf{Top}: a lightweight gate receives the hidden state from the frozen VLA backbone and decides per frame whether to activate language generation or directly output waypoints. \textbf{Bottom}: gate training pipeline. Labels are derived from route-level success rate comparisons and refined via frame-level reconstruction. The visual encoder and LLM backbone remain frozen \raisebox{-0.15em}{\includegraphics[height=1.0em]{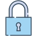}}, while only a lightweight MLP gate is trainable \raisebox{-0.15em}{\includegraphics[height=1.0em]{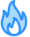}}.}
  \vspace{-2pt}
  \label{fig:method}
\end{figure*}

\paragraph{Setup.} We study SimLingo \cite{renz2025simlingo}, a VLA driving model that generates natural language reasoning before predicting actions. By skipping language generation, the model can also predict actions directly from its internal representations. We evaluate both configurations on all 44 scenario categories of Bench2Drive \cite{jia2024bench2drive}, running repeated experiments per route. Figure~\ref{fig:when_analysis} shows the per-scenario results. We provide analysis under additional settings in Appendix~\ref{sec:appendix_extra_settings}.

\paragraph{Language Does Not Always Help.} As shown in Figure~\ref{fig:when_analysis}, language generation only matters on a small fraction of routes, but where it matters, the impact on driving performance is substantial, either improving or degrading it by a large margin. On the majority of routes, language has no measurable effect yet still incurs full generation cost. Intuitively, if we can detect when language is needed and skip it otherwise, the model could achieve both better driving performance and faster inference.

\paragraph{Room for Improvement.} To quantify the potential gains, we construct a route-level oracle that picks the better-performing configuration for each route. Even with this coarse-grained selection, the oracle already reaches 78.4\% success rate, revealing more than 10\% room for improvement over the default VLA. Since the oracle only makes a single choice per route while finer per-frame selection could further improve performance, this estimate is conservative. The large gap confirms that a lightweight selection mechanism can unlock substantial performance gains without any model retraining, motivating the gate design in Section~\ref{sec:method}.

\section{Method}\label{sec:method}
In this section, we present BLUE, which uses pretrained VLA hidden states to predict per frame whether to activate language generation. Figure~\ref{fig:method} shows the overall framework.

\paragraph{Hidden States Encode Language Necessity.}
To predict when language generation is needed, we look for a signal within the model itself. We find that a simple logistic regression trained on the VLA's last-layer hidden states can distinguish frames where language helps from those where it does not, without relying on any external features. This shows that the pretrained hidden states potentially already encode a language-utility signal, providing a basis for building a lightweight gate.

\paragraph{Data Collection and Labeling.}
We run both language mode and direct action mode on training routes through repeated experiments, collecting the last-layer hidden state $\mathbf{h} \in \mathbb{R}^{d}$ at the final token position for each frame. 
Details of data splits are provided in Appendix~\ref{sec:appendix_data_splits}, and additional labeling details are provided in Appendix~\ref{sec:appendix_labeling_details}.
This position corresponds to the model's representation right before language or waypoint generation begins, and is shared by both modes to ensure feature alignment.
See more details of data splits and labeling in Appendix~\ref{sec:appendix_frame_labels}.
Each route $r$ is first assigned a binary label based on the success rate gap:
\begin{equation}\label{eq:label_route}
    y_r = \mathbbm{1}\!\Big[\frac{1}{|\mathcal{S}|}\sum_{s \in \mathcal{S}} \big(\mathrm{SR}_{\text{lang}}^{(r,s)} - \mathrm{SR}_{\text{direct}}^{(r,s)}\big) > \tau\Big],
\end{equation}
where $\mathcal{S}$ is the set of random seeds, $\mathrm{SR}$ denotes success rate, and $\tau{=}10\%$ is a margin threshold. This design defaults to the faster direct action mode unless language shows a clear advantage.

\begin{table*}[t!]
\centering
\renewcommand{\arraystretch}{1.08}
\resizebox{\textwidth}{!}{%
\begin{tabular}{@{}llcclc cc@{}}
\toprule
\multirow{2}{*}{\textbf{Method}} & \multicolumn{5}{c}{\textbf{Details}} & \multicolumn{2}{c}{\textbf{Metrics}} \\ \cmidrule(lr){2-6} \cmidrule(lr){7-8}
& Expert & Camera & LiDAR & Labels & T-Param. & SR (\%) $\uparrow$ & DS $\uparrow$ \\ \midrule
UniAD-Base \cite{hu2023planning} & Think2Drive & 6$\times$ & \xmark & O,M,S & $\geq$\,59\,M & 16.36 & 45.81 \\
TF++ \cite{zimmerlin2024hidden} & PDM-Lite & 1$\times$ & \cmark & O,M,S,D & $\geq$\,39\,M & 67.27 & 84.21 \\
MomAD \cite{song2025don} & Think2Drive & 6$\times$ & \xmark & O,M & $\geq$\,25\,M & 18.11 & 47.91 \\
DriveTrans \cite{jia2025drivetransformer} & Think2Drive & 6$\times$ & \xmark & O,M & $\approx$\,646\,M & 35.01 & 63.46 \\
Hydra-NeXt \cite{li2025hydra} & Think2Drive & 2$\times$ & \xmark & - & $\geq$\,25\,M & 50.00 & 73.86 \\
DiffusionDrive \cite{liao2025diffusiondrive} & - & 3$\times$ & \cmark & O,S & $\approx$\,60\,M & 52.72 & 77.68 \\
ORION \cite{fu2025orion} & Think2Drive & 6$\times$ & \xmark & O,M,L & $\geq$\,300\,M & 54.62 & 77.74 \\
AutoVLA \cite{zhou2026autovla} & PDM-Lite & 1$\times$ & \xmark & L & $\geq$\,1.5\,B & 57.73 & 78.84 \\
SimLingo \cite{renz2025simlingo} & PDM-Lite & 1$\times$ & \xmark & L & $\geq$\,300\,M & 67.27 & 85.07 \\
HiP-AD \cite{tang2025hip} & Think2Drive & 6$\times$ & \xmark & O,M,D & $\approx$\,97\,M & 69.09 & 86.77 \\
ReCogDrive \cite{li2025recogdrive} & Think2Drive & 6$\times$ & \xmark & L & $\geq$\,2\,B & 45.45 & 71.36 \\
GeRo \cite{yasarla2026generative} & Think2Drive & 6$\times$ & \xmark & O,M,L & $\geq$\,3\,B & 60.10 & 81.90 \\
DeLL \cite{du2026deconfounded} & PDM-Lite & 1$\times$ & \cmark & O,S & $\geq$\,38\,M & 68.63 & 86.86 \\
R2SE \cite{liu2026reinforced} & PDM-Lite & 1$\times$ & \cmark & O,M,S,D & $\geq$\,39\,M & 69.54 & 86.28 \\
AutoMoT \cite{huang2026automot} & PDM-Lite & 1$\times$ & \cmark & - & $\approx$\,1.6\,B & 70.00 & 87.34 \\
BevAD \cite{holtz2026matters} & PDM-Lite & 6$\times$ & \xmark & O & $\geq$\,25\,M & 72.73 & 88.11 \\
CriticVLA \cite{yang2026judge} & PDM-Lite & 1$\times$ & \xmark & L & $\geq$\,300\,M & 73.33 & 88.02 \\
TakeVLA \cite{gao2026learning} & PDM-Lite & 1$\times$ & \xmark & L & $\geq$\,300\,M & 73.73 & 89.72 \\
\addlinespace[2pt]
\midrule
\rowcolor{bbb!15}\textbf{BLUE (Ours)} & PDM-Lite & 1$\times$ & \xmark & L & \textbf{0.11\,M} & \textbf{76.18{\scriptsize$\pm$0.64}} & \textbf{90.58{\scriptsize$\pm$0.12}} \\
$\Delta$ vs. SimLingo & - & - & - & - & - & +8.91 & +5.51 \\ \bottomrule
\end{tabular}%
}
\caption{Results on Bench2Drive. BLUE achieves the highest closed-loop success rate (SR) and driving score (DS), with large margins over its SimLingo backbone. T-Param.\ reports trainable parameters; we use published values ($\approx$) where available and conservative lower bounds ($\geq$) derived from the minimum size of trained components. BLUE trains only a 0.11M gate while keeping the VLA backbone frozen. Notably, BLUE surpasses methods that employ multi-camera setups, LiDAR, or dense auxiliary labels (O: 3D object detection, M: map, S: semantic segmentation, D: depth, L: language), using only front-view camera with language annotations. See more results in Appendix \ref{sec:appendix_full_b2d}.}
\label{tab:driving_results}
\vspace{-5pt}
\end{table*}

We construct training labels at two granularities. In route-level labeling, every frame on route $r$ shares the same label $y_r$. In frame-level labeling, we further refine routes that benefit from language ($y_r{=}1$) by marking only the critical regions $\mathcal{C}_r$ where language mode and direct action mode differ the most. We identify these regions from spatial patterns of behavior divergence. See Appendix~\ref{sec:appendix_frame_labels} for details. The frame-level label is:

\begin{equation}\label{eq:label_frame}
    y_{r,t} = \mathbbm{1}\!\big[\Delta\overline{\mathrm{SR}}_r > \tau\big] \cdot \mathbbm{1}\!\big[\mathbf{x}_t \in \mathcal{C}_r\big],
\end{equation}

where $\Delta\overline{\mathrm{SR}}_r {=} \frac{1}{|\mathcal{S}|}\sum_{s} (\mathrm{SR}_{\text{lang}}^{(r,s)} {-} \mathrm{SR}_{\text{direct}}^{(r,s)})$ is the cross-seed language advantage of route $r$, and $\mathbf{x}_t {\in} \mathbb{R}^2$ is the spatial coordinate of frame $t$. The first indicator selects language-beneficial routes, while the second restricts positive labels to critical segments within those routes. We mix samples from both labeling granularities during training so that the gate learns both route-level preference and frame-level activation.
To address temporal redundancy (e.g., near-identical features when the vehicle is stopped), we group consecutive frames with cosine similarity above $0.99$ into redundant segments and downsample each segment of length $L$ to $\max(2, \lceil\sqrt{L}\rceil)$ representative samples.

\paragraph{Gate Design.}
The gate is a single-hidden-layer MLP with negligible parameter count, trained with binary cross-entropy loss. This small design is sufficient for mapping pretrained hidden states to a binary gating decision, while reducing overfitting risk and keeping inference overhead negligible.

\paragraph{On-demand language use.}
The trained gate is integrated into the VLA inference pipeline with negligible overhead. At each frame, the model computes $\mathbf{h}$ through a forward pass shared by both modes. The gate outputs a score $p(\mathbf{h})$: if it exceeds a threshold $\theta$, the model proceeds with language generation; otherwise, it produces actions directly. The threshold $\theta$ governs how selectively the gate triggers language generation at inference time.

\begin{table*}[t!]
\centering
\resizebox{\textwidth}{!}{%
\begin{tabular}{lcccccccc}
\toprule
\multirow{2}{*}{\textbf{Method}} & \multicolumn{2}{c}{\textbf{Details}} & \multicolumn{6}{c}{\textbf{Multi-Ability Success Rate (\%)}} \\ \cmidrule(lr){2-3} \cmidrule(lr){4-9}
& Camera & LiDAR & Merge $\uparrow$ & Overtake $\uparrow$ & EmBrake $\uparrow$ & GiveWay $\uparrow$ & TSign $\uparrow$ & Mean $\uparrow$ \\
\midrule
UniAD-Base \cite{hu2023planning} & 6$\times$ & \xmark & 14.10 & 17.78 & 21.67 & 10.00 & 14.21 & 15.55 \\
TF++ \cite{zimmerlin2024hidden} & 1$\times$ & \cmark & 58.75 & 57.77 & 83.33 & 40.00 & 82.11 & 64.39 \\
DriveTrans \cite{jia2025drivetransformer} & 6$\times$ & \xmark & 17.57 & 35.00 & 48.36 & 40.00 & 52.10 & 38.60 \\
Hydra-NeXt \cite{li2025hydra} & 2$\times$ & \xmark & 40.00 & 64.44 & 61.67 & 50.00 & 50.00 & 53.22 \\
DiffusionDrive \cite{liao2025diffusiondrive} & 3$\times$ & \cmark & 50.63 & 26.67 & 68.33 & 50.00 & 76.32 & 54.38 \\
ORION \cite{fu2025orion} & 6$\times$ & \xmark & 25.00 & 71.11 & 78.33 & 30.00 & 69.15 & 54.72 \\
HiP-AD \cite{tang2025hip} & 6$\times$ & \xmark & 50.00 & 84.44 & 83.33 & 40.00 & 72.10 & 65.98 \\
SimLingo \cite{renz2025simlingo} & 1$\times$ & \xmark & 53.78 & 67.41 & 81.67 & 50.00 & 77.20 & 66.01 \\
ReCogDrive \cite{li2025recogdrive} & 6$\times$ & \xmark & 29.73 & 20.00 & 69.09 & 20.00 & 71.34 & 42.03 \\
GeRo \cite{yasarla2026generative} & 6$\times$ & \xmark & 40.06 & 78.24 & 87.32 & 50.00 & 76.83 & 66.49 \\
R2SE \cite{liu2026reinforced} & 1$\times$ & \cmark & 53.33 & 61.25 & 90.00 & 50.00 & 84.21 & 67.76 \\
DeLL \cite{du2026deconfounded} & 1$\times$ & \cmark & 61.25 & 62.22 & 80.00 & 60.00 & 81.05 & 68.90 \\
TakeVLA \cite{gao2026learning} & 1$\times$ & \xmark & 63.64 & 64.44 & 91.67 & 50.00 & 85.48 & 71.05 \\
CriticVLA \cite{yang2026judge} & 1$\times$ & \xmark & 61.28 & 76.30 & 88.33 & 50.00 & 81.06 & 71.39 \\
BevAD \cite{holtz2026matters} & 6$\times$ & \xmark & 71.67 & 74.07 & 75.56 & 76.67 & 75.44 & 74.68 \\ 
\midrule
\rowcolor{bbb!15}BLUE (ours) & 1$\times$ & \xmark & 61.44{\scriptsize$\pm$1.33} & 80.00{\scriptsize$\pm$1.81} & 93.27{\scriptsize$\pm$1.33} & 50.00{\scriptsize$\pm$0.00} & 84.74{\scriptsize$\pm$0.00} & 73.89{\scriptsize$\pm$0.14} \\
$\Delta$ vs. SimLingo & - & - & +7.66 & +12.59 & +11.60 & +0.00 & +7.54 & +7.88 \\ \bottomrule
\end{tabular}%
}
\caption{Multi-ability results on Bench2Drive. Mean denotes the average success rate over the five driving skills. While using only a single front-view camera and no LiDAR, BLUE achieves the second-best mean result and remains close to the best method, which uses six cameras. See additional results in Appendix~\ref{sec:appendix_additional_results}.}
\label{tab:multi_ability}
\end{table*}

\section{Experiments}\label{sec:experiments}
In this section, we evaluate BLUE on two established closed-loop driving benchmarks and compare against a wide range of published methods.

\paragraph{Setup.}
We apply BLUE to SimLingo \cite{renz2025simlingo} with a gate threshold of $\theta{=}0.66$. The gate uses a hidden dimension of 128 and is trained on approximately 400 routes sampled from SimLingo's training set. We evaluate on two benchmarks: Bench2Drive \cite{jia2024bench2drive}, a multi-scenario benchmark covering 220 routes across 44 scenario categories in CARLA \cite{dosovitskiy2017carla}, and Longest6 v2 \cite{carla_garage}, a long-horizon benchmark that evaluates sustained driving quality through driving score, route completion, and infraction score. We compare against 26 published methods spanning end-to-end and VLA approaches. All BLUE results are averaged over 3 random seeds. More details on data splits, benchmarks, baselines, implementation, and additional results are provided in Appendix~\ref{sec:appendix_experimental_details} and~\ref{sec:appendix_additional_results}.

\paragraph{Closed-Loop Results on Bench2Drive.}
Table~\ref{tab:driving_results} presents the main comparison on Bench2Drive. BLUE achieves the highest success rate and driving score among all compared methods, improving both metrics over its backbone SimLingo by a large margin. 
Notably, BLUE trains only a 0.11M-parameter gate on a frozen backbone, uses a single front-view camera without LiDAR, and requires only language annotations. Despite this minimal setup, it surpasses methods that employ six cameras, LiDAR, dense auxiliary labels, or orders-of-magnitude more trainable parameters.
Table~\ref{tab:multi_ability} further reports the multi-ability breakdown. BLUE ranks second in mean ability score, close to the best method that relies on six cameras, with clear improvements in overtaking and emergency braking.

\paragraph{Closed-Loop Results on Longest6 v2.}
Table~\ref{tab:longest} presents results on Longest6 v2, a long-horizon benchmark that evaluates sustained driving quality. BLUE achieves the highest driving score and route completion among all compared methods, while requiring substantially fewer GPU hours. The improvement in route completion suggests that our BLUE helps maintain robust driving over long distances, where errors from unnecessary language generation would otherwise compound.

\begin{table}[t]
\centering
\small
\renewcommand{\arraystretch}{1.1}
\resizebox{\linewidth}{!}{%
\begin{tabular}{@{}lcccc@{}}
\toprule
\textbf{Method} & \textbf{DS} $\uparrow$ & \textbf{RC} $\uparrow$ & \textbf{IS} $\uparrow$ & \textbf{Time} $\downarrow$ \\
\midrule
HiP-AD \cite{tang2025hip} & 7 & 56 & - & - \\
TF++ \cite{zimmerlin2024hidden} & 23 & 71 & - & - \\
SimLingo \cite{renz2025simlingo} & 22 & 70 & 0.38 & 119h \\
CriticVLA \cite{yang2026judge} & 34 & 66 & \textbf{0.55} & 193h \\
\midrule
\rowcolor{bbb!15}BLUE (ours) & \textbf{36} & \textbf{84} & 0.43 & \textbf{56h} \\
$\Delta$ vs. SimLingo & +14 & +14 & +0.05 & -63h \\
\bottomrule
\end{tabular}%
}
\caption{Closed-loop results on Longest6 v2. DS: driving score, RC: route completion, IS: infraction score. Time: total A100 GPU hours to evaluate all routes.}
\label{tab:longest}
\end{table}

\paragraph{Inference Efficiency.}
Since the gate skips language generation on most frames, BLUE runs substantially faster than the full language mode. The gate itself adds negligible overhead, as it is a single-hidden-layer MLP applied to the already-computed hidden state. As shown in Table~\ref{tab:efficiency}, BLUE achieves a 2.54$\times$ speedup over SimLingo with the lowest latency among all compared methods.
We provide additional efficiency analysis in the next section.

\begin{table}[t]
\centering
\small
\renewcommand{\arraystretch}{1.1}
\setlength{\tabcolsep}{4pt}
\begin{tabular*}{\linewidth}{@{\extracolsep{\fill}}lccc@{}}
\toprule
\textbf{Method} & \textbf{Speed Ratio} $\uparrow$ & \textbf{FPS} $\uparrow$ & \textbf{Latency (ms)} $\downarrow$ \\
\midrule
HiP-AD & 0.0625 & 1.25 & 800.3 \\
SimLingo & 0.0358 & 0.72 & 1396.6 \\
CriticVLA & 0.0146 & 0.29 & 3424.7 \\
\midrule
\rowcolor{bbb!15}BLUE (ours) & \textbf{0.0910} & \textbf{1.82} & \textbf{549.5} \\
$\Delta$ vs. SimLingo & +154.2\% & +154.2\% & -60.7\% \\
\bottomrule
\end{tabular*}
\caption{Inference efficiency comparison among representative driving models. Higher speed ratio and FPS are better, while lower latency is better.}
\label{tab:efficiency}
\end{table}

\section{Analysis and Ablation Study}\label{sec:analysis}

We now analyze the language gate from multiple angles: its activation behavior, generalizability to other models, and sensitivity to design choices.

\begin{mdframed}[backgroundcolor=bbb!8]
\begin{minipage}{\linewidth}
\noindent (1) What activation pattern does the gate learn?
\vspace{-12pt}
\end{minipage}
\end{mdframed}

\begin{figure*}[t]
  \centering
  \includegraphics[width=0.98\textwidth]{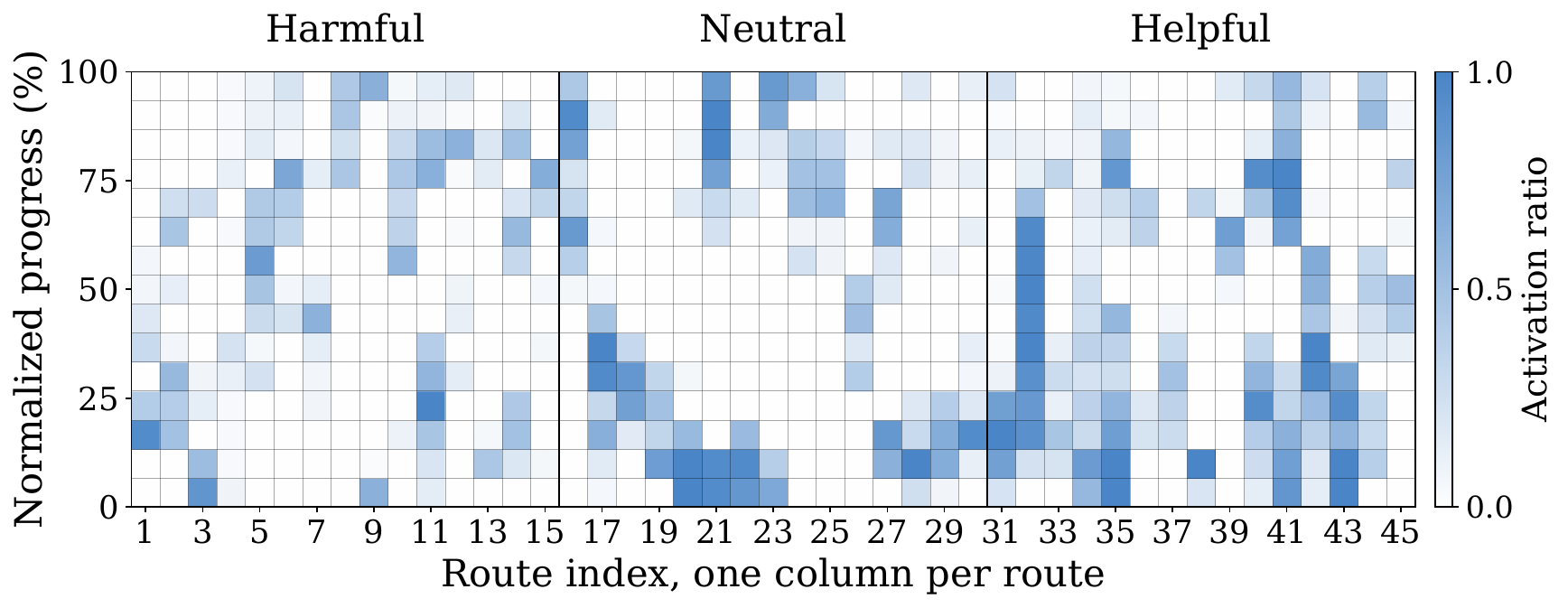}
  \vspace{-8pt}
  \caption{Language activation pattern learned by the gate. Each column is one route, grouped by the route-level language benefit measured in Section~\ref{sec:when}. The color encodes the fraction of frames where language generation is activated at each progress bin. The gate activates language sparingly and in contiguous segments.}
  \label{fig:temporal}
\end{figure*}

We visualize the gate's frame-level decisions across evaluation routes in Figure~\ref{fig:temporal}. The gate skips language generation on most frames, yet BLUE still outperforms the VLA backbone by a large margin, as shown in Section~\ref{sec:experiments}. When it does activate language, the activations form contiguous segments rather than scattering across isolated frames, suggesting that the gate captures temporally coherent patterns from the hidden states rather than producing noisy frame-level fluctuations.

\begin{mdframed}[backgroundcolor=bbb!8]
\begin{minipage}{\linewidth}
\noindent (2) Can BLUE be applied to other models?
\end{minipage}
\end{mdframed}

The main experiments use SimLingo as the backbone. To examine whether BLUE transfers to other architectures, we apply the same pipeline to CriticVLA \cite{yang2026judge}, a VLA model with a different language integration design. As shown in Table~\ref{tab:criticvla}, BLUE improves CriticVLA across all metrics on both Bench2Drive and Longest6 v2, with particularly notable gains in route completion on Longest6 v2. The resulting system also surpasses all other listed baselines. This suggests that the language-utility signal in hidden states is not unique to SimLingo, and that the gate mechanism can benefit other language-centric VLA models. Full comparison results are in Appendix~\ref{sec:appendix_additional_results}.

\begin{table}[t]
\centering
\small
\renewcommand{\arraystretch}{1.1}
\setlength{\tabcolsep}{3.5pt}
\resizebox{\linewidth}{!}{%
\begin{tabular}{@{}lcccc@{}}
\toprule
\multirow{2}{*}{\textbf{Method}} & \multicolumn{2}{c}{\textbf{Bench2Drive}} & \multicolumn{2}{c}{\textbf{Longest6 v2}} \\ \cmidrule(lr){2-3} \cmidrule(lr){4-5}
& SR (\%) $\uparrow$ & DS $\uparrow$ & DS $\uparrow$ & RC $\uparrow$ \\
\midrule
TF++ \cite{zimmerlin2024hidden} & 67.27 & 84.21 & 23 & 71 \\
HiP-AD \cite{tang2025hip} & 69.09 & 86.77 & 7 & 56 \\
SimLingo \cite{renz2025simlingo} & 67.27 & 85.07 & 22 & 70 \\
CriticVLA \cite{yang2026judge} & 73.33 & 88.02 & 34 & 66 \\
\midrule
\rowcolor{bbb!15}BLUE (CriticVLA) & \textbf{76.04} & \textbf{90.37} & \textbf{36.2} & \textbf{80.6} \\
$\Delta$ vs. CriticVLA & +2.71 & +2.35 & +2.2 & +14.6 \\
\bottomrule
\end{tabular}%
}
\caption{BLUE applied to CriticVLA, compared with representative methods on Bench2Drive and Longest. Full comparison results are in Appendix~\ref{sec:appendix_additional_results}.}
\label{tab:criticvla}
\end{table}

\begin{mdframed}[backgroundcolor=bbb!8]
\begin{minipage}{\linewidth}
\noindent (3) Is the gate transferable across models?
\end{minipage}
\end{mdframed}

Since BLUE trains a separate gate for each model, a natural question is whether a gate learned on one model can be directly reused on another without retraining. We test this by swapping the trained gates between SimLingo and CriticVLA. As shown in Table~\ref{tab:transfer}, the matched gate consistently outperforms the transferred gate by a clear margin. This indicates that when language helps is inherently tied to the model itself rather than to the driving scenario. Different models internalize language utility in distinct ways within their hidden states, so each model should train its own gate.
Appendix~\ref{sec:appendix_extra_settings} further analyzes why each model requires its own gate and its retraining cost.

\begin{table}[t]
\centering
\small
\renewcommand{\arraystretch}{1.2}
\setlength{\tabcolsep}{4pt}
\begin{tabular*}{\linewidth}{@{\extracolsep{\fill}}lcc@{}}
\toprule
\textbf{Configuration} & \textbf{SR (\%)} $\uparrow$ & \textbf{DS} $\uparrow$ \\
\midrule
SimLingo \cite{renz2025simlingo} & 67.27{\scriptsize$\pm$2.11} & 85.07{\scriptsize$\pm$0.95} \\
SimLingo + CriticVLA gate  & 71.59{\scriptsize$\pm$0.96} & 89.23{\scriptsize$\pm$0.21} \\
SimLingo + SimLingo gate & \textbf{76.18}{\scriptsize$\pm$0.64} & \textbf{90.58}{\scriptsize$\pm$0.12} \\
\midrule
CriticVLA \cite{yang2026judge} & 73.33{\scriptsize$\pm$0.27} & 88.02{\scriptsize$\pm$0.17} \\
CriticVLA + SimLingo gate & 73.11{\scriptsize$\pm$1.36} & 88.90{\scriptsize$\pm$0.53} \\
CriticVLA + CriticVLA gate & \textbf{76.04}{\scriptsize$\pm$0.38} & \textbf{90.37}{\scriptsize$\pm$0.14} \\
\bottomrule
\end{tabular*}
\caption{Cross-model transfer of the learned language gate on Bench2Drive. Each transferred gate is evaluated on a model different from the one used for training.}
\label{tab:transfer}
\end{table}

\begin{mdframed}[backgroundcolor=bbb!8]
\begin{minipage}{\linewidth}
\noindent (4) Are rule-based methods sufficient for predicting when language is needed?
\end{minipage}
\end{mdframed}

We ask whether simple driving signals can replace the learned gate. We design three rule-based gates that activate language when a kinematic feature exceeds a threshold: vehicle speed, acceleration magnitude, or steering angle. See Appendix~\ref{sec:appendix_baseline_gates} for construction details. As shown in Table~\ref{tab:gate_baselines}, regardless of which feature or threshold is used, all rule-based gates fall far short of BLUE and offer only marginal gains over single-mode baselines. The random gate performs even worse. This result is expected: kinematic features reflect only the vehicle's current motion and do not provide enough information to predict whether language will help. The VLA's hidden states, which jointly encode perceptual and contextual information, are a much stronger signal for this prediction.

\begin{table}[t]
\centering
\small
\renewcommand{\arraystretch}{1.1}
\setlength{\tabcolsep}{4pt}
\resizebox{\linewidth}{!}{%
\begin{tabular}{@{}lccc@{}}
\toprule
\textbf{Gate Method} & \textbf{SR (\%)} $\uparrow$ & \textbf{DS} $\uparrow$ & \textbf{Lang. (\%)} \\
\midrule
Speed-based gate & 70.97{\scriptsize$\pm$1.21} & 88.71{\scriptsize$\pm$0.63} & 55.50{\scriptsize$\pm$1.21} \\
Speed-based gate & 71.81{\scriptsize$\pm$0.54} & 89.91{\scriptsize$\pm$0.88} & 30.21{\scriptsize$\pm$0.91} \\
Acceleration-based gate & 70.08{\scriptsize$\pm$0.43} & 88.33{\scriptsize$\pm$0.43} & 49.12{\scriptsize$\pm$0.87} \\
Steering-based gate & 70.71{\scriptsize$\pm$0.97} & 89.38{\scriptsize$\pm$0.22} & 7.94{\scriptsize$\pm$0.02} \\
\midrule
Complexity-based gate & 70.98{\scriptsize$\pm$0.80} & 87.12{\scriptsize$\pm$1.75} & 53.61{\scriptsize$\pm$1.08} \\
Complexity-based gate & 71.40{\scriptsize$\pm$2.26} & 88.02{\scriptsize$\pm$0.85} & 17.15{\scriptsize$\pm$0.58} \\
\midrule
Random gate & 67.42{\scriptsize$\pm$0.01} & 86.66{\scriptsize$\pm$1.18} & 79.90{\scriptsize$\pm$0.16} \\
Random gate & 70.96{\scriptsize$\pm$0.71} & 88.36{\scriptsize$\pm$0.16} & 50.07{\scriptsize$\pm$0.41} \\
Random gate & 70.01{\scriptsize$\pm$1.53} & 87.10{\scriptsize$\pm$1.58} & 20.15{\scriptsize$\pm$0.32} \\
\midrule
\rowcolor{bbb!15}BLUE (ours) & \textbf{76.18}{\scriptsize$\pm$0.64} & \textbf{90.58}{\scriptsize$\pm$0.12} & 21.44{\scriptsize$\pm$0.66} \\
\bottomrule
\end{tabular}
}
\caption{Comparison of alternative gating strategies on Bench2Drive. Kinematic gates activate language when a motion feature exceeds a threshold; the complexity gate activates language on complex routes. Lang. denotes the percentage of frames with language activation.}
\label{tab:gate_baselines}
\end{table}

\begin{mdframed}[backgroundcolor=bbb!8]
\begin{minipage}{\linewidth}
\noindent (5) Is scenario complexity sufficient for predicting when language is needed?
\end{minipage}
\end{mdframed}

An intuitive hypothesis is that language generation is needed in complex scenarios and can be skipped in simple ones. To test this, we compute a composite complexity score for each training route based on structured features including the number of sub-scenarios, weather severity, traffic flow density, and opposing vehicle interactions. Routes above a threshold are labeled complex and activate language generation, while the rest skip it and predict actions directly. See Appendix~\ref{sec:appendix_baseline_gates} for more details. As shown in Table~\ref{tab:gate_baselines}, the complexity-based gate performs comparably to the kinematic-based gates and remains far below BLUE. This indicates that whether language generation helps is not determined by scenario complexity alone, and frame-level hidden states capture dynamics that coarse route-level labels cannot provide. We discuss why complexity-based gate fall short in Appendix~\ref{sec:appendix_extra_settings}.

\begin{mdframed}[backgroundcolor=bbb!8]
\begin{minipage}{\linewidth}
\noindent (6) Does BLUE improve inference efficiency?
\vspace{-12pt}
\end{minipage}
\end{mdframed}

\begin{figure}[t]
  \centering
  \includegraphics[width=0.99\linewidth]{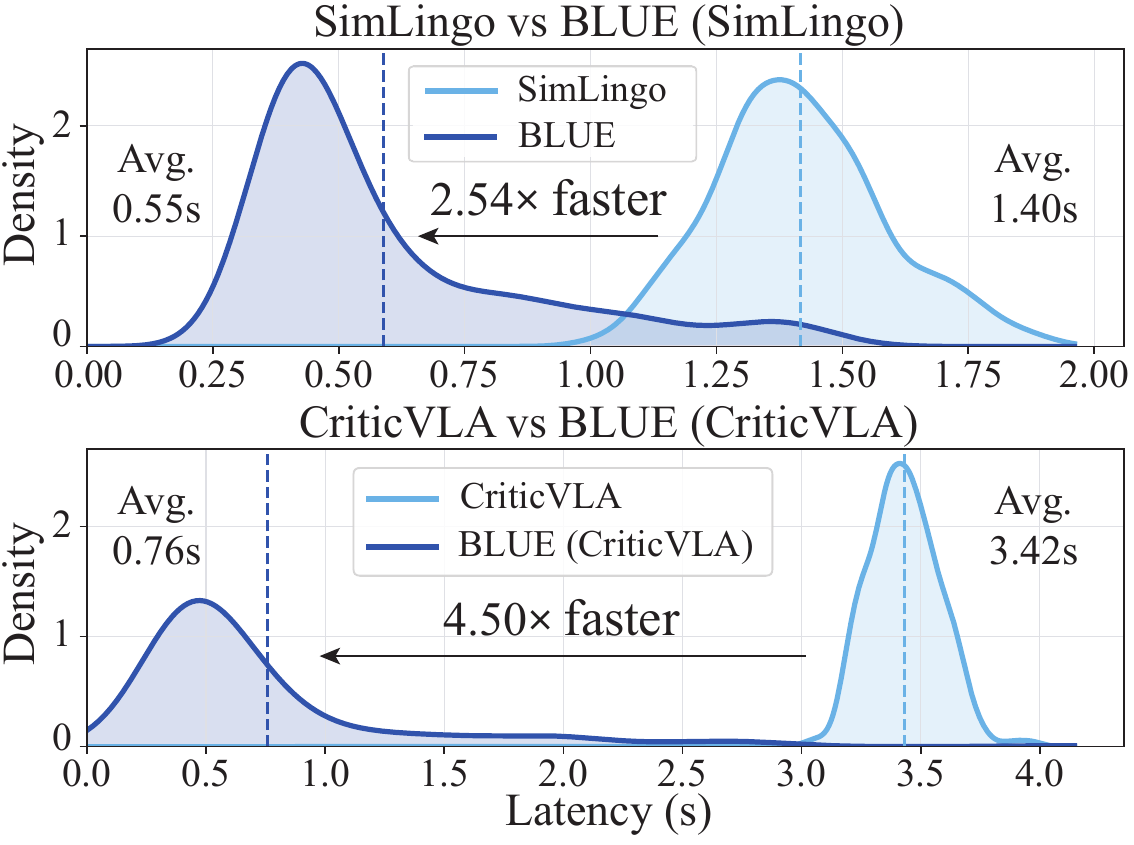}
  \vspace{-20pt}
  \caption{Distribution of mean per-frame latency across Bench2Drive evaluation routes. Dashed lines indicate overall means. BLUE substantially reduces latency for both backbones, delivering 2.54$\times$ and 4.50$\times$ speedup.}
  \label{fig:efficiency}
\end{figure}

Beyond driving performance, BLUE also improves inference efficiency. Since the gate skips language generation on most frames and the gate itself is a single-hidden-layer MLP with negligible overhead, BLUE reduces the per-frame latency substantially. Table~\ref{tab:efficiency} reports the aggregate statistics: BLUE on SimLingo achieves 2.54$\times$ speedup and reduces mean latency from 1.40\,s to 0.55\,s. The gain extends to CriticVLA, where BLUE achieves 4.50$\times$ speedup and lowers the mean latency from 3.42\,s to 0.76\,s. Figure~\ref{fig:efficiency} further shows the per-route latency distributions across all evaluation routes. For both backbones, BLUE shifts the entire distribution toward lower latency. These results confirm that BLUE simultaneously improves both driving quality and inference efficiency.

\begin{mdframed}[backgroundcolor=bbb!8]
\begin{minipage}{\linewidth}
\noindent (7) How sensitive is the gate to the threshold?
\end{minipage}
\end{mdframed}

\begin{figure}[t]
  \centering
  \includegraphics[width=0.99\linewidth]{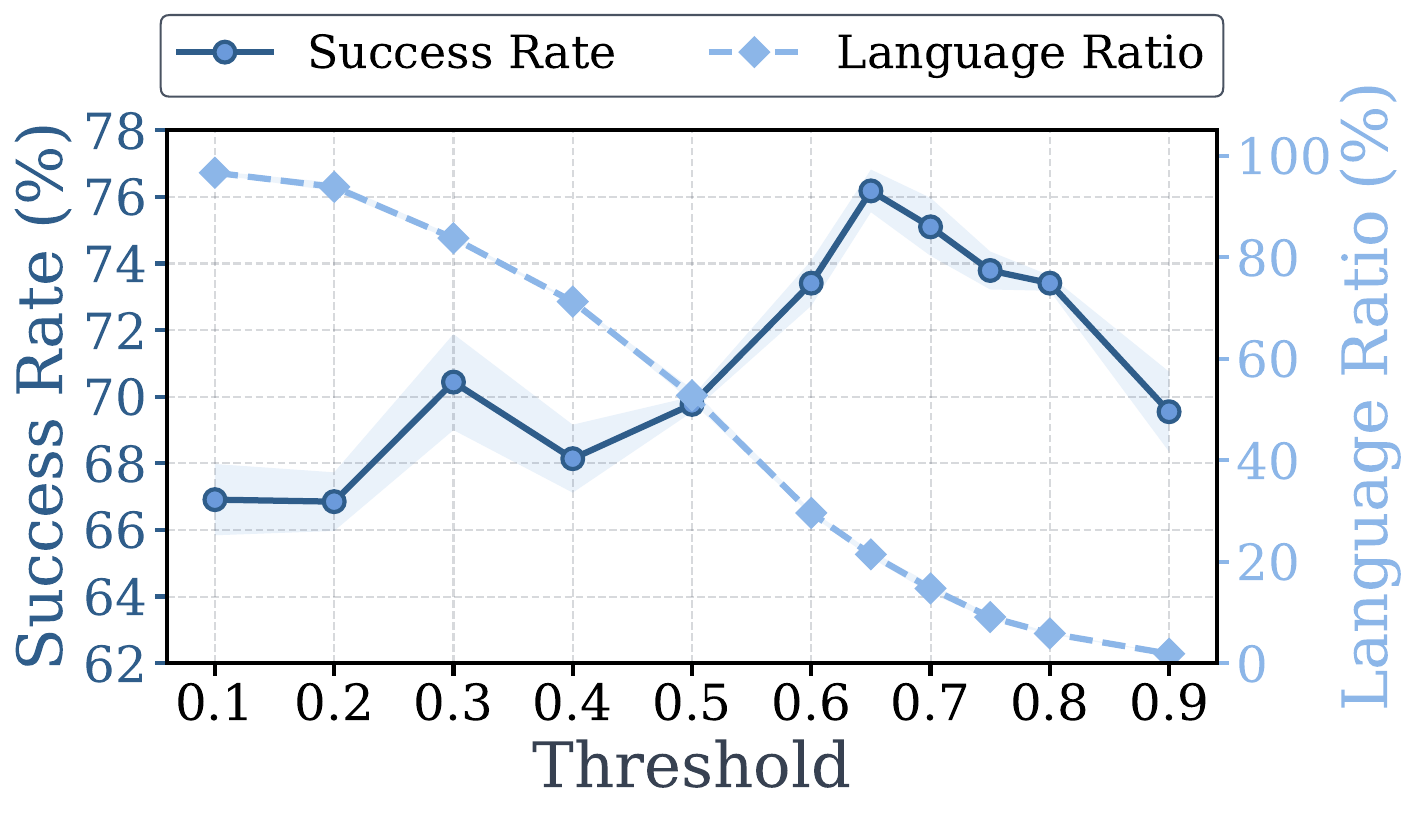}
  \vspace{-20pt}
  \caption{Effect of threshold $\theta$ on language activation ratio and success rate. Language usage decreases monotonically with $\theta$, while $\theta \in [0.6, 0.8]$ yields strong SR.}
  \label{fig:threshold}
\end{figure}

The gate threshold $\theta$ controls how readily the model activates language generation. We select $\theta$ without tuning based on a simple observation: the effect of language on each route falls into three natural categories, helpful, neutral, and harmful, which partition the gate output range $[0, 1]$ into three equal intervals. We place $\theta$ at the boundary between neutral and helpful, yielding $\theta{=}0.66$ directly from this categorization.
Figure~\ref{fig:threshold} shows the effect of varying $\theta$ across the full range. Language activation ratio decreases monotonically as $\theta$ increases, confirming that the gate produces well-calibrated scores. SR peaks near our chosen value, and thresholds from 0.6 to 0.8 all achieve good results. When $\theta$ is very low, the model generates language reasoning on nearly every frame before acting, and SR reduces to 66.91\%. When $\theta$ is very high, the model skips language and produces actions directly at every frame, yielding SR~=~69.55\%. BLUE at $\theta{=}0.66$ achieves 76.18\% SR, surpassing these two settings by a large margin.

\begin{mdframed}[backgroundcolor=bbb!8]
\begin{minipage}{\linewidth}
\noindent (8) How does training data size affect the gate?
\vspace{-12pt}
\end{minipage}
\end{mdframed}

\begin{figure}[t]
  \centering
  \includegraphics[width=0.99\linewidth]{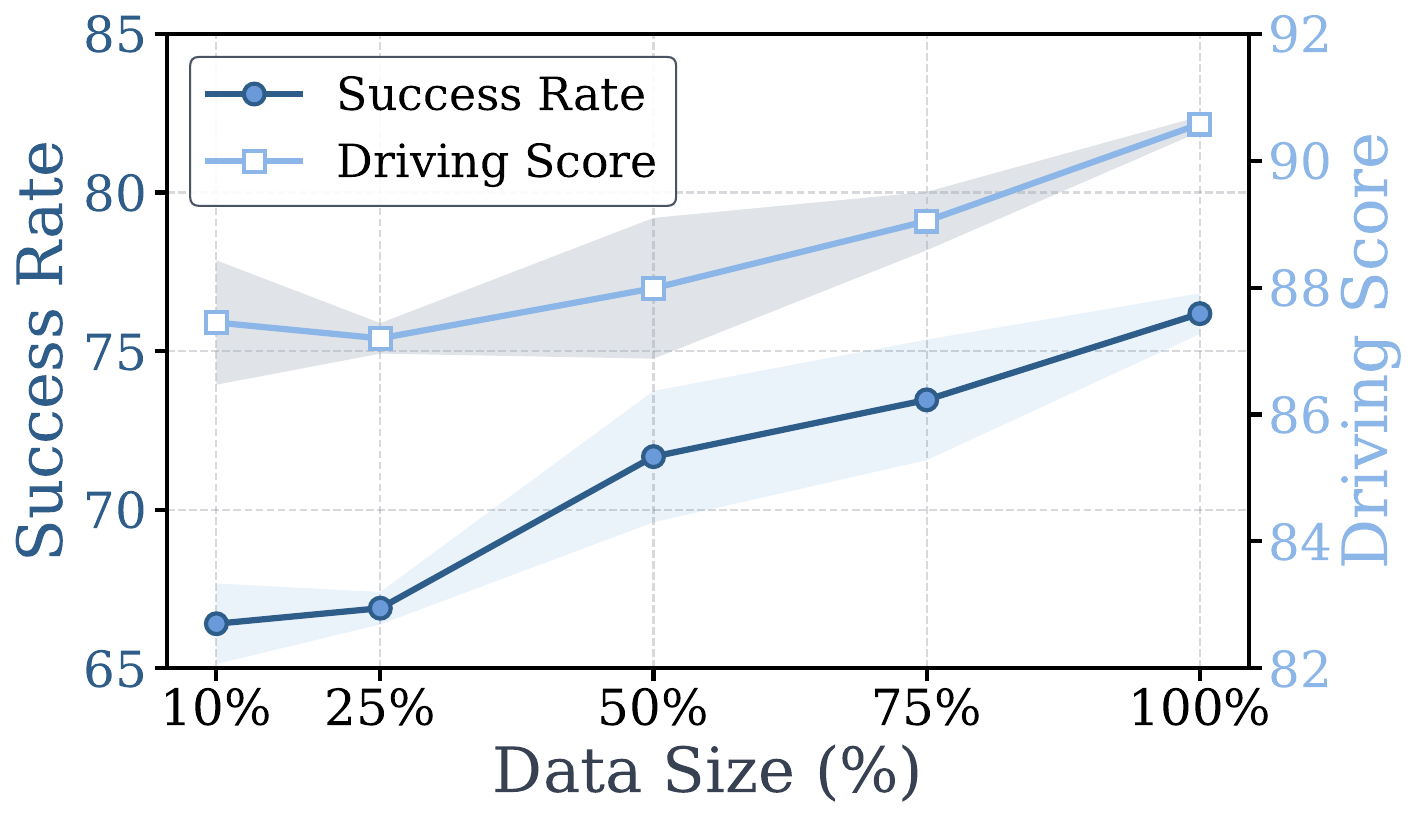}
  \vspace{-20pt}
  \caption{Effect of training data size on gate performance. Both SR and DS improve quickly with more training data and then approach saturation.}
  \label{fig:data_size}
\end{figure}

We collect approximately 400 training routes from the SimLingo training set, vary the proportion used from 10\% to 100\%, and evaluate closed-loop driving on Bench2Drive. Training and evaluation routes have no overlap. We detail the data splits in Appendix~\ref{sec:appendix_data_splits}. As shown in Figure~\ref{fig:data_size}, both SR and DS improve steadily as training data increases. With only half of the available routes, the gate already surpasses the SimLingo backbone by a clear margin, indicating that the language-utility signal encoded in hidden states is learnable from moderate amounts of data. Performance continues to improve with additional training routes, and variance across seeds decreases, reflecting more stable gate decisions. We use the full training set in all other experiments for best results.

\begin{mdframed}[backgroundcolor=bbb!8]
\begin{minipage}{\linewidth}
\noindent (9) How does gate design affect performance?
\vspace{-12pt}
\end{minipage}
\end{mdframed}

We vary the gate hidden dimension and dropout setting while fixing all other factors. As shown in Table~\ref{tab:gate_arch}, all configurations achieve strong performance, and increasing gate capacity does not bring clear improvement. Among all variants, dropout provides a noticeable benefit. We adopt the smaller gate with dropout as the default, since a smaller capacity combined with regularization better prevents overfitting to training routes. 
This choice is made purely based on the principle of minimizing overfitting risk, not from test-set tuning.

\begin{table}[t]
\centering
\small
\renewcommand{\arraystretch}{1.1}
\setlength{\tabcolsep}{4pt}
\begin{tabular}{@{}cccc@{}}
\toprule
\textbf{Hidden Dim} & \textbf{Dropout} & \textbf{Success Rate} $\uparrow$ & \textbf{Driving Score} $\uparrow$ \\
\midrule
128 & No & 74.67{\scriptsize$\pm$0.69} & 90.25{\scriptsize$\pm$0.20} \\
\rowcolor{bbb!15}128 & Yes & \textbf{76.18}{\scriptsize$\pm$0.64} & \textbf{90.58}{\scriptsize$\pm$0.12} \\
256 & No & 74.62{\scriptsize$\pm$0.33} & 89.97{\scriptsize$\pm$0.36} \\
256 & Yes & 75.19{\scriptsize$\pm$1.56} & 89.70{\scriptsize$\pm$1.15} \\
\bottomrule
\end{tabular}
\caption{Effect of gate hidden dimension and dropout on Bench2Drive. All settings use the same training data, labels, and inference threshold.}
\label{tab:gate_arch}
\end{table}

\begin{mdframed}[backgroundcolor=bbb!8]
\begin{minipage}{\linewidth}
\noindent (10) Is the impact of language use consistent across experimental settings?
\end{minipage}
\end{mdframed}
In Section~\ref{sec:when}, we observed a performance gap and complementarity between driving with and without language generation on SimLingo. Here we test whether this pattern persists across different experimental settings by varying annotation granularity from normal to brief, switching annotation language from English to Chinese, and replacing the backbone from SimLingo to CriticVLA.
As shown in Table~\ref{tab:annotation}, the same pattern holds across all settings: language improves driving on only a minority of routes, actively degrades it on another subset, and has no clear effect on the majority. 
This consistent complementarity across annotation settings and backbones supports applying BLUE to different VLA models. 
We provide statistical testing details for each setting in Appendix~\ref{sec:appendix_extra_settings}.

\begin{table}[t]
\centering
\small
\renewcommand{\arraystretch}{1.1}
\resizebox{\linewidth}{!}{%
\begin{tabular}{@{}llccc@{}}
\toprule
\textbf{Category} & \textbf{Setting} & \textbf{Helpful} & \textbf{Neutral} & \textbf{Harmful} \\
\midrule
Original & SimLingo & 14.5\% & 61.8\% & 23.6\% \\
Granularity & Brief & 21.8\% & 52.7\% & 25.5\% \\
Language & Chinese & 18.6\% & 58.2\% & 23.2\% \\
Model & CriticVLA & 19.5\% & 52.3\% & 28.2\% \\
\bottomrule
\end{tabular}
}
\caption{Language usefulness across different settings. Helpful, neutral, and harmful denote the proportion of routes where language helps, has no effect, or hurts.}
\label{tab:annotation}
\end{table}

\begin{table*}[t]
\centering
\small
\renewcommand{\arraystretch}{1.08}
\setlength{\tabcolsep}{2.8pt}
\arrayrulecolor{black}
\resizebox{\textwidth}{!}{%
\begin{tabular}{@{}l|cccc|c|ccc|ccc@{}}
\toprule
\multirow{3}{*}{\textbf{Method}}
& \multirow{3}{*}{\rotatebox[origin=c]{90}{RGB}}
& \multirow{3}{*}{\rotatebox[origin=c]{90}{LiDAR}}
& \multirow{3}{*}{\rotatebox[origin=c]{90}{Privileged}}
& \multirow{3}{*}{\rotatebox[origin=c]{90}{Learned}}
& \multicolumn{1}{c|}{\textbf{Bench2Drive}}
& \multicolumn{6}{c}{\textbf{Fail2Drive}} \\
\cmidrule(lr){6-6}\cmidrule(lr){7-12}
& & & & &
\multirow{2}{*}{DS $\uparrow$}
& \multicolumn{3}{c|}{\textbf{In-Distribution}}
& \multicolumn{3}{c}{\textbf{Generalization}} \\
\cmidrule(lr){7-9}\cmidrule(lr){10-12}
& & & & &
& DS $\uparrow$ & SR (\%) $\uparrow$ & HM $\uparrow$
& DS $\uparrow$ & SR (\%) $\uparrow$ & HM $\uparrow$ \\
\midrule

TCP~\citeyearpar{wu2022trajectory}
& \cmark & \xmark & \xmark & \cmark
& 59.9
& 24.7 & 39.1 & 30.3
& 24.5 {\scriptsize(-0.8\%)}
& 31.4 {\scriptsize(-19.7\%)}
& 27.5 {\scriptsize(-9.1\%)} \\

UniAD~\citeyearpar{hu2023planning}
& \cmark & \xmark & \xmark & \cmark
& 45.8
& 47.5 & 36.3 & 41.2
& 44.0 {\scriptsize(-7.4\%)}
& 27.6 {\scriptsize(-24.0\%)}
& 33.9 {\scriptsize(-17.6\%)} \\

ORION~\citeyearpar{fu2025orion}
& \cmark & \xmark & \xmark & \cmark
& 77.8
& 53.0 & 52.0 & 52.5
& 51.2 {\scriptsize(-3.4\%)}
& 46.0 {\scriptsize(-11.5\%)}
& 48.5 {\scriptsize(-7.7\%)} \\

HiP-AD~\citeyearpar{tang2025hip}
& \cmark & \xmark & \xmark & \cmark
& 86.8
& 74.1 & 70.7 & 72.4
& 67.1 {\scriptsize(-9.4\%)}
& 56.7 {\scriptsize(-19.8\%)}
& 61.5 {\scriptsize(-15.1\%)} \\

SimLingo~\citeyearpar{renz2025simlingo}
& \cmark & \xmark & \xmark & \cmark
& 85.1
& 82.6 & 79.3 & 80.9
& 71.7 {\scriptsize(-13.2\%)}
& 55.0 {\scriptsize(-30.6\%)}
& 62.2 {\scriptsize(-23.1\%)} \\

AlignDrive~\citeyearpar{wu2026aligndrive}
& \cmark & \xmark & \xmark & \cmark
& 89.1
& 74.7 & 75.7 & 75.2
& 68.6 {\scriptsize(-8.1\%)}
& 62.7 {\scriptsize(-17.2\%)}
& 65.5 {\scriptsize(-12.9\%)} \\

BevAD~\citeyearpar{holtz2026matters}
& \cmark & \xmark & \xmark & \cmark
& 88.1
& 87.4 & 83.3 & 85.3
& 82.3 {\scriptsize\,(-5.8\%)}
& 68.7 {\scriptsize\,(-17.6\%)}
& 74.9 {\scriptsize\,(-12.2\%)} \\

\rowcolor{bbb!15}
BLUE (ours)
& \cmark & \xmark & \xmark & \cmark
& 90.6
& 85.7 & 84.0 & 84.8
& 73.9 {\scriptsize(-13.8\%)}
& 59.0 {\scriptsize(-29.8\%)}
& 65.6 {\scriptsize(-22.7\%)} \\

\midrule

TF++~\citeyearpar{zimmerlin2024hidden}
& \cmark & \cmark & \xmark & \cmark
& 84.2
& 83.3 & 78.5 & 80.8
& 75.4 {\scriptsize(-9.5\%)}
& 61.1 {\scriptsize(-22.2\%)}
& 67.5 {\scriptsize(-16.5\%)} \\

TFv6~\citeyearpar{nguyen2026lead}
& \cmark & \cmark & \xmark & \cmark
& 95.0
& 90.2 & 93.3 & 91.7
& 79.5 {\scriptsize(-11.8\%)}
& 70.7 {\scriptsize(-24.3\%)}
& 74.8 {\scriptsize(-18.4\%)} \\

BridgeDrive~\citeyearpar{liu2025bridgedrive}
& \cmark & \cmark & \xmark & \cmark
& 96.3
& 91.6 & 95.0 & 93.3
& 81.9 {\scriptsize(-10.5\%)}
& 75.0 {\scriptsize(-21.1\%)}
& 78.3 {\scriptsize(-16.0\%)} \\

\midrule

PlanT 2.0~\citeyearpar{gerstenecker2025plant}
& -- & -- & \cmark & \cmark
& 92.4
& 87.8 & 85.0 & 86.4
& 73.3 {\scriptsize(-16.5\%)}
& 58.0 {\scriptsize(-31.8\%)}
& 64.8 {\scriptsize(-25.0\%)} \\

\midrule

PDMLite-F2D
& -- & -- & \cmark & \xmark
& 97.0
& 95.6 & 97.0 & 96.3
& 94.0 {\scriptsize(-1.7\%)}
& 95.3 {\scriptsize(-1.8\%)}
& 94.6 {\scriptsize(-1.7\%)} \\

\bottomrule
\end{tabular}%
}
\caption{Results on Fail2Drive. In-Distribution covers standard CARLA scenarios, while Generalization covers out-of-distribution long-tail scenarios. HM is the harmonic mean of DS and SR. Parentheses report the relative change from In-Distribution to Generalization. Our results are averaged over three random seeds.}
\label{tab:fail2drive}
\end{table*}

\begin{mdframed}[backgroundcolor=bbb!8]
\begin{minipage}{\linewidth}
\noindent (11) Can BLUE handle long-tail scenarios?
\end{minipage}
\end{mdframed}

To assess generalization to unseen long-tail scenarios, we evaluate BLUE on Fail2Drive~\cite{gerstenecker2026fail2drive}, a closed-loop benchmark featuring rare cases such as animal crossings and fully blocked roads. As shown in Table~\ref{tab:fail2drive}, BLUE outperforms its SimLingo backbone on every metric in both the in-distribution and generalization settings, while achieving the highest in-distribution success rate among camera-only methods. In the challenging generalization test, BLUE improves DS from 71.7 to 73.9, SR from 55.0\% to 59.0\%, and HM from 62.2 to 65.6, confirming that its gains over SimLingo persist under long-tail conditions.

\begin{table*}[t]
\centering
\small
\setlength{\tabcolsep}{7pt}
\renewcommand{\arraystretch}{1.15}
\resizebox{\textwidth}{!}{%
\begin{tabular}{@{}lcccccccc@{}}
\toprule
\textbf{Method} & \textbf{Lang. Ratio} & \textbf{EP} $\uparrow$ & \textbf{NC} $\uparrow$ & \textbf{DAC} $\uparrow$ & \textbf{DDC} $\uparrow$ & \textbf{TTC} $\uparrow$ & \textbf{Comfort} $\uparrow$ & \textbf{PDMS} $\uparrow$ \\
\midrule
ReCogDrive~\citeyearpar{li2025recogdrive} & 100\% & 79.08 & 97.64 & 92.99 & 97.80 & 92.88 & \textbf{99.99} & 84.13 \\
ReCogDrive~\citeyearpar{li2025recogdrive} & 0\% & 80.53 & 98.19 & 94.14 & 97.83 & 94.09 & 99.98 & 85.98 \\
\midrule
\rowcolor{bbb!15}
BLUE (ReCogDrive) & 38.27\% & \textbf{81.35} & \textbf{98.50} & \textbf{94.77} & \textbf{97.88} & \textbf{94.84} & \textbf{99.99} & \textbf{87.00} \\
\bottomrule
\end{tabular}%
}
\caption{Results on NAVSIM. All settings use the same ReCogDrive backbone and differ only in language use. BLUE outperforms both ReCogDrive baselines, demonstrating its generalizability to real-world driving data.}
\label{tab:navsim}
\end{table*}

\begin{mdframed}[backgroundcolor=bbb!8]
\begin{minipage}{\linewidth}
\noindent (12) Does BLUE improve real-world driving?
\end{minipage}
\end{mdframed}

To assess whether BLUE extends beyond simulation, we integrate it with ReCogDrive~\cite{li2025recogdrive} and evaluate it on NAVSIM~\cite{dauner2024navsim}, a benchmark built from real-world driving data. ReCogDrive supports two ways to produce a trajectory. With language generation, it autoregressively generates the trajectory as text and converts it into numerical waypoints for execution. When language generation is skipped, it generates no tokens and instead passes the VLM hidden state to a planner trained by imitation learning on the NAVSIM training set, which predicts the trajectory directly. BLUE uses the frozen VLM hidden state to choose between these two options at each frame. We train the gate on 90\% of the NAVSIM training set and use the remaining 10\% to select its decision threshold. As shown in Table~\ref{tab:navsim}, BLUE outperforms both fixed strategies, showing that selective language use remains effective on real-world driving data. However, evaluation on recorded real-world data is not equivalent to deployment on real vehicles. Such deployment would require closed-loop testing under sensor noise, distribution shifts, and safety-critical conditions, as discussed in Appendix~\ref{sec:appendix_potential_risk}.

\section{Conclusion}\label{sec:conclusion}

We present BLUE, a minimal method for better language use in VLA driving. 
We reveal that generated language helps in some situations, hurts in others, and is unnecessary most of the time. 
Based on this observation, BLUE trains a 0.11M-parameter gate on frozen VLA hidden states to decide when to generate language. 
It achieves 76.2\% success rate on Bench2Drive, 36 driving score on Longest6 v2, setting new state of the art while delivering 2.54$\times$ inference speedup. 
Results across three VLA backbones and four benchmarks further demonstrate the broad applicability of BLUE.
These results suggest that better language use in VLA driving does not come from generating more language, but from generating language only when it improves driving.

\clearpage

\section*{Limitations}
\label{sec:limitations}
Despite its effectiveness, BLUE has two limitations.
First, BLUE introduces uneven per-frame latency, as frames that skip language generation run faster than those that generate language. However, this is not unique to BLUE. Language-generating VLA systems inherently exhibit variable per-frame latency because the number of output tokens differs across frames. Since the gate adds negligible overhead, BLUE barely increases the maximum per-frame latency compared with the original VLA, while substantially reducing the average.
Second, BLUE requires training a separate gate for each backbone. However, the gate is a lightweight single-layer MLP and the VLA backbone remains entirely frozen, so the adaptation cost is low. Training labels are a natural byproduct of routine driving evaluation and require no additional human annotation or reward engineering. Applying BLUE to a new backbone is therefore considerably cheaper than retraining or modifying the backbone itself.

\section*{Acknowledgments}
AI assistants were used in this work exclusively for English language polishing and assisting with code implementation. All research ideas, experimental designs, analyses, and scientific conclusions are original contributions of the authors. The authors reviewed all LLM-assisted outputs and take full responsibility for the final content of this paper.

\bibliography{custom}

\clearpage
\appendix

\etocdepthtag{appendix}
\etocsettagdepth{main}{none}
\etocsettagdepth{appendix}{subsubsection}

\newcommand{\tocdot}{\nobreak\leaders\hbox to 0.8em{\hss.\hss}\hfill}




\etocsetstyle{section}
  {\par}
  {\addvspace{0.30em}}
  {\noindent \textbf{\etocnumber{} \etocname} \tocdot \etocpage \par}
  {}

\etocsetstyle{subsection}
  {}
  {\addvspace{0.15em}}
  {\noindent \hspace{0.5em} \etocnumber{} \etocname \tocdot \etocpage \par}
  {}

\etocsetstyle{subsubsection}
  {}
  {\addvspace{0.13em}}
  {\noindent \hspace{1.5em} \etocnumber{} \etocname \tocdot \etocpage \par}
  {}

{
  \setlength{\parskip}{2.25pt}
  \tableofcontents
}

\newpage

\section{The Novelty and Contribution of BLUE}

In this section, we summarize the main novelty of BLUE and explain how does this work contribute to the NLP community. Detailed comparisons with related methods are provided in Section~\ref{sec:appendix_rw_comparison}.

\subsection{Novelty of BLUE}

BLUE is related to recent work on language-augmented driving, efficient reasoning, and adaptive computation. We highlight five aspects that distinguish BLUE, while deferring detailed method-by-method comparisons to Section~\ref{sec:appendix_rw_comparison}.

\paragraph{Systematic diagnosis of language utility.}
Prior work on adaptive reasoning focuses on improving, accelerating, or selectively invoking language generation, but none has systematically quantified how generated language affects closed-loop driving outcomes. We conduct large-scale closed-loop evaluations on hundreds of routes and statistically categorize each route as language-helpful, language-neutral, or language-harmful. The results show that language improves driving on only a minority of routes, actively degrades it on another subset, and has no clear effect on the remaining majority. This finding offers a new perspective for the community: selective language use in VLA driving should not only reduce unnecessary computation but also actively prevent harmful language generation.

\paragraph{Hidden-state language-utility signal.}
Existing methods rely on reinforcement learning rewards, scene complexity features, or architectural modifications to learn when to reason, all of which require retraining or modifying the VLA backbone. BLUE discovers that the pretrained hidden states of a frozen VLA potentially already encode whether language generation will benefit driving at the current frame. This means the gating decision does not require redesigning the model or additional training of the backbone. The gate simply reads out a signal that already exists inside the pretrained representations, keeping the original VLA unchanged and avoiding the cost of backbone-level retraining.

\paragraph{Low-cost label collection.}
Existing adaptive-reasoning methods typically require constructing reward functions or curating specialized training data. BLUE derives its training labels by simply comparing route success when the VLA generates language versus when it predicts actions directly. This process requires no human annotation and no reward engineering. In autonomous driving, deploying or validating a model already involves collecting driving outcomes under different configurations. BLUE reuses these routine evaluation results as supervision, making label collection a natural byproduct of the standard development pipeline rather than an additional burden.

\paragraph{State-of-the-art results with full reproducibility.}
Despite training only a 0.11M-parameter MLP gate on a frozen VLA backbone, BLUE achieves state-of-the-art closed-loop driving results on both Bench2Drive and Longest6 v2 while delivering 2.54$\times$ inference speedup. Many concurrent driving methods remain closed-source, making their results difficult to reproduce or build upon. We fully release the code, training data, model checkpoints, and evaluation logs, providing the community with a fully reproducible baseline for future research on better language use in VLA driving.

\paragraph{Toward better language use in vision and robotics.}
As language models are increasingly used in visual and robotic systems, language generation must improve downstream behavior under latency constraints. Existing approaches mainly optimize the amount or speed of generation. BLUE demonstrates a complementary principle: better language use requires deciding when to generate, not only how to generate. In embodied settings, unnecessary language can slow the system and steer actions in harmful ways. Maximizing language's net positive impact therefore requires generation decisions that adapt to each input. This insight extends to any deployed system where language serves as an intermediate computation.

\section{Related Work}
\label{sec:appendix_related_work}

BLUE sits at the intersection of end-to-end autonomous driving, adaptive reasoning, and driving evaluation. We first review end-to-end driving methods with a focus on language-augmented models (\S\ref{sec:appendix_rw_ad_methods}), then discuss efficient reasoning techniques in both VLA driving (\S\ref{sec:appendix_rw_adaptive_vla}) and LLMs (\S\ref{sec:appendix_rw_adaptive_reasoning}), followed by the benchmarks used for evaluation (\S\ref{sec:appendix_rw_benchmarks}). Finally, we provide a detailed comparison between BLUE and existing methods (\S\ref{sec:appendix_rw_comparison}).

\subsection{End-to-End Autonomous Driving}
\label{sec:appendix_rw_ad_methods}
Autonomous driving has evolved from modular pipelines that chain perception, prediction, and planning into end-to-end models that directly map sensor inputs to driving actions~\cite{chen2024end, hu2023planning, wu2022trajectory, li2025end}. Within this paradigm, diverse architectural designs have emerged, including multi-modal transformer fusion~\cite{prakash2021multi, chitta2022transfuser, shao2023safety}, vectorized scene representation for planning~\cite{hu2023planning, jiang2023vad, li2024ego, jia2025drivetransformer}, learning-based planner supervision~\cite{dauner2023parting, li2024hydra, wu2026aligndrive, li2025learning}, and diffusion-based or generative trajectory prediction~\cite{liao2025diffusiondrive, zheng2024genad, liu2025bridgedrive}. A growing line of work further integrates natural language into the driving loop through vision-language models and vision-language-action models~\cite{zhang2024vision, zhou2022learning, zhong2024let, huang2025causality, wang2026unifying}, which generate scene descriptions or reasoning chains as intermediate representations before producing control outputs~\cite{sima2024drivelm, tian2024drivevlm, shao2024lmdrive, hwang2024emma, jiang2024senna, tang2026hermes, ling2026agent}. Complementary directions include world models for predictive simulation~\cite{hu2023gaia, gao2024vista, liu2026unidwm} and reinforcement learning for driving policy optimization~\cite{kiran2021deep, sallab2017deep, wang2018deep}. 

\subsection{Adaptive Reasoning in Driving VLA}
\label{sec:appendix_rw_adaptive_vla}

Recent VLA driving models typically generate language reasoning before predicting actions, but language generation introduces significant inference overhead. Several concurrent methods have explored strategies to reduce this cost. One line of work trains VLA models to select reasoning depth adaptively. AutoVLA~\cite{zhou2026autovla} builds a unified autoregressive VLA with physical action tokenization and applies supervised fine-tuning followed by GRPO-based reinforcement fine-tuning to reduce reasoning in straightforward scenarios. AdaThinkDrive~\cite{luo2025adathinkdrive} trains a dual-mode think and non-think policy through supervised learning and GRPO with an adaptive think reward. Another line of work introduces multi-system architectures. DE-Driver \cite{xie2026deliberation} designs a dual-expert VLA with a scene-aware router that dispatches inputs to a reactive or deliberative expert. SAMoE-VLA~\cite{you2026samoe} extends this idea with scene-adaptive mixture-of-experts routing, and FASIONAD~\cite{qian2024fasionad} adopts a dual-system framework where a fast system handles routine navigation while a slow system reasons from visual prompts and provides feedback in challenging situations. A related direction seeks alternative reasoning representations. FutureX~\cite{lin2025futurex} uses an auto-think switch to choose between instant planning and latent world-model reasoning, invoking CoT-guided latent rollout only when additional reasoning is needed. DynVLA~\cite{shang2026dynvla} introduces a dynamics chain-of-thought that compresses future world evolution into compact dynamics tokens. From the efficiency perspective, FastDriveCoT~\cite{gu2026accelerating} applies parallel decoding to structured chain-of-thought templates, and Reasoning-VLA~\cite{zhang2025reasoning} replaces autoregressive action decoding with learnable action queries for parallel trajectory generation. We provide a detailed comparison between BLUE and these methods in \S\ref{sec:appendix_rw_comparison_vla}.

\subsection{Efficient Reasoning in LLMs}
\label{sec:appendix_rw_adaptive_reasoning}
BLUE detects when language generation is needed in VLA driving. This question is related to efficient reasoning in NLP, where recent methods control the length of chain-of-thought through RL with length rewards or SFT on variable-length reasoning data.

\subsubsection{RL with Length Reward Design}
Early RL training for reasoning models focuses on accuracy rewards, which often leads to unnecessarily verbose chains of thought. To address this, recent methods incorporate length-based penalties into the reward function so that shorter correct answers receive higher scores. Kimi k1.5 \cite{team2025kimi} adds a length penalty to its policy optimization to control long reasoning activations. O1-Pruner \cite{luo2025o1} introduces a length-harmonizing reward with PPO, optimizing the ratio of reasoning lengths between a reference model and the student. L1 \cite{aggarwal2025l1} appends explicit token-budget constraints to the input before applying GRPO \cite{shao2024deepseekmath}. Demystifying Long CoT \cite{yeo2025demystifying} proposes a cosine-shaped reward that penalizes excessive length while stabilizing training. DAST \cite{shen2025dast} constructs length-preference data and trains with SimPO \cite{meng2024simpo} to adapt reasoning depth to problem difficulty. Arora et al.\ \cite{arora2026training} condition length rewards on correctness, assigning higher scores to shorter correct solutions. Other representative efforts include AdaptThink \cite{zhang2025adaptthink}, S-GRPO \cite{dai2026s}, and ConciseRL \cite{dumitru2025conciserl}. These methods share the goal of producing concise reasoning in text-based question answering and mathematics. BLUE addresses a related but distinct setting: instead of shortening textual reasoning, it decides whether to activate language generation at all in an embodied VLA driving model.

\begin{table*}[t]
\centering
\small
\renewcommand{\arraystretch}{1.36}
\setlength{\tabcolsep}{3pt}
\resizebox{\textwidth}{!}{%
\begin{tabular}{@{}lllcl@{}}
\toprule
\textbf{Method} & \textbf{What is Adapted} & \textbf{Core Mechanism} & \textbf{Frozen} & \textbf{Supervision} \\
\midrule
AutoVLA~\cite{zhou2026autovla} & Reasoning depth & Action token. + SFT + GRPO & \xmark & RL reward \\
AdaThinkDrive~\cite{luo2025adathinkdrive} & Think / non-think & Dual-mode SFT + GRPO & \xmark & Adaptive think reward \\
DE-Driver~\cite{xie2026deliberation} & Reactive / delib. expert & Dual-expert + scene router & \xmark & Scene-aware routing \\
SAMoE-VLA~\cite{you2026samoe} & Expert assignment & Scene-adaptive MoE & \xmark & Scene features \\
FASIONAD~\cite{qian2024fasionad} & Fast / slow system & Dual-system + VLM feedback & \xmark & Confidence score \\
FutureX~\cite{lin2025futurex} & Instant / latent thinking & Auto-think + world model & \xmark & World-model rollout \\
DynVLA~\cite{shang2026dynvla} & Text / dynamics CoT & Dynamics token prediction & \xmark & SFT on dyn. tokens \\
FastDriveCoT~\cite{gu2026accelerating} & CoT decoding speed & Parallel structured decoding & \xmark & Template structure \\
Reasoning-VLA~\cite{zhang2025reasoning} & Action decoding & Action queries + parallel gen. & \xmark & SFT \\
\midrule
\rowcolor{bbb!15}
BLUE (ours) & Language gen. (on / off) & Hidden-state gate (0.11M) & \cmark & Language utility \\
\bottomrule
\end{tabular}%
}
\caption{Comparison of BLUE with related adaptive reasoning methods for VLA driving. \textbf{Frozen} indicates whether the original VLA backbone weights remain frozen. BLUE is the only method that performs post-hoc language gating on a frozen VLA, with labels derived from closed-loop driving outcomes.}
\label{tab:vla_comparison}
\end{table*}

\subsubsection{SFT with Variable-Length CoT}
\label{sec:appendix_rw_sft_variable_cot}
Beyond RL, supervised fine-tuning on curated variable-length reasoning data provides another route to efficient reasoning. These methods differ mainly in how the short chains are constructed. In the post-reasoning approach, Distilling System 2 into System 1 \cite{yu2024distilling} removes the reasoning process entirely and distills only the final answer. C3oT \cite{kang2025c3ot} uses GPT-4 as a compressor to shorten reasoning while retaining key information. TokenSkip \cite{xia2025tokenskip} estimates the semantic importance of each reasoning segment and removes low-importance tokens. In the during-reasoning approach, Learn to Skip \cite{liu2024can} first creates concise solutions by manually merging or removing steps, then trains the model to intrinsically skip steps during inference. Token-Budget \cite{han2025token} uses binary search to find the optimal token budget and trains the model to follow it. Self-Training \cite{munkhbat2025self} samples multiple reasoning paths and selects the shortest correct one as training data. CoT-Valve \cite{ma2025cot} progressively mixes parameters of long-reasoning and non-reasoning models to generate variable-length training data. Other related works include ReCUT \cite{jin2025recut}, ConCISE \cite{qiao2025concise} and NCoTS \cite{ling2026neural}. Like RL-based methods, these approaches focus on compressing textual reasoning chains in language tasks. BLUE operates at a coarser granularity: rather than shortening the generated language, it uses a lightweight gate to decide whether language generation should be activated for a given frame.

\subsection{Autonomous Driving Benchmarks}
\label{sec:appendix_rw_benchmarks}
Evaluating autonomous driving models falls into two broad categories depending on whether the model's own outputs influence future states. Open-loop benchmarks, including nuScenes~\cite{caesar2020nuscenes}, Waymo Open~\cite{sun2020scalability}, and Argoverse 2~\cite{wilson2023argoverse}, measure prediction quality against recorded ground truth but do not let the model's actions affect subsequent observations. As a result, prediction errors that would compound during actual driving remain hidden in open-loop evaluation. Closed-loop benchmarks fill this gap by requiring the model to drive full routes inside a simulator or on replayed logs, where cumulative effects on driving outcomes such as route completion and collision avoidance can be directly measured. On the simulator side, CARLA~\cite{dosovitskiy2017carla} provides a configurable urban environment that supports diverse traffic and weather conditions. Built on CARLA, Bench2Drive~\cite{jia2024bench2drive} defines 220 routes across 44 scenario categories and evaluates models on multiple metrics such as success rate and driving score, covering five distinct driving skills. Longest6 v2~\cite{carla_garage} contains 36 long routes of 1--2 km each and evaluates sustained driving quality over extended distances. Fail2Drive~\cite{gerstenecker2026fail2drive} evaluates generalization across 100 standard routes and 100 rare, challenging long-tail routes, such as those involving animal crossings or fully blocked roads. On the log-replay side, nuPlan~\cite{caesar2021nuplan}, WayMax~\cite{gulino2023waymax}, NavSim~\cite{dauner2024navsim}, and OpenAD~\cite{li2026open} replay real-world driving logs with reactive or non-reactive traffic agents. We evaluate BLUE on three closed-loop benchmarks: Bench2Drive, Longest6 v2, and Fail2Drive. Together, they provide complementary evaluations of multi-scenario driving, long-horizon driving, and generalization to challenging long-tail scenarios. To further assess BLUE on real-world driving data, we also conduct experiments on NAVSIM~\cite{dauner2024navsim}. Experiments across all four benchmarks broadly validate the effectiveness of BLUE.

\subsection{Comparison with Related Methods}
\label{sec:appendix_rw_comparison}
We now compare BLUE with adaptive reasoning methods in VLA driving (\S\ref{sec:appendix_rw_comparison_vla}) and with efficient reasoning methods in LLMs (\S\ref{sec:appendix_rw_comparison_llm}).

\subsubsection{Adaptive Reasoning in VLA}
\label{sec:appendix_rw_comparison_vla}
Section~\ref{sec:appendix_rw_adaptive_vla} surveys the methods covered here. Table~\ref{tab:vla_comparison} provides a structured comparison of BLUE with these methods.
As shown in Table~\ref{tab:vla_comparison}, BLUE differs from existing methods in four aspects. 

First, existing adaptive reasoning methods focus on learning when to activate longer or shorter reasoning, but none of them systematically examines how generated language affects closed-loop driving performance. BLUE fills this gap by conducting extensive evaluations and quantitatively categorizing language impact into helpful, neutral, and harmful cases, offering insights into when and why language generation should be selectively applied. 

Second, all prior methods require modifying the VLA backbone or introducing new architectural components, whereas BLUE keeps the VLA entirely frozen and trains only a 0.11M-parameter gate. The minimal parameter count means the gate can be trained in minutes on a single GPU, adds negligible inference overhead, and avoids overfitting on limited calibration data, making BLUE a lightweight plug-in applicable to many existing VLA model without retraining. 

Third, existing methods rely on reinforcement learning with custom reward functions, scene-level features, or new training objectives, all of which demand additional supervision and pipeline changes. BLUE instead discovers that pretrained VLA hidden states potentially already encode whether language generation will benefit driving, and the gate simply reads out this existing signal. The training labels are collected automatically by running the VLA and comparing driving outcomes, requiring no human annotation, no reward engineering, and minimal additional cost, as the data collection can be integrated into the routine road testing that deployed driving models already undergo.

Fourth, existing adaptive VLA methods such as AutoVLA and AdaThinkDrive \cite{zhou2026autovla, luo2025adathinkdrive} base their reasoning decisions on scenario complexity, activating longer reasoning in difficult scenes and reducing it in simple ones. However, as demonstrated in Section~\ref{sec:analysis}, complexity-based gates perform comparably to simple kinematic heuristics and remain far below BLUE. The underlying reason, analyzed in detail in Appendix~\ref{sec:appendix_extra_settings}, is that whether language helps depends not only on the scenario but also on how a specific model processes and utilizes language. The same scenario can be language-helpful under one model yet language-harmful under another, making complexity an unreliable proxy. BLUE addresses this limitation by conditioning the gating decision on model-specific hidden states, which jointly encode perceptual context and the model's internal readiness to benefit from language generation.

\subsubsection{Efficient Reasoning in LLM}
\label{sec:appendix_rw_comparison_llm}

Section~\ref{sec:appendix_rw_adaptive_reasoning} surveys RL-based and SFT-based methods for controlling reasoning length in LLMs. BLUE shares the motivation of avoiding unnecessary reasoning but differs in three ways. First, LLM methods \cite{team2025kimi, luo2025o1, huang2025routereval, yeo2025demystifying} control how long a reasoning chain should be, producing shorter or longer traces depending on problem difficulty. BLUE makes a binary decision: whether language generation should be activated at all for a given driving frame. This coarser formulation suits VLA driving, where the two inference modes produce qualitatively different action distributions rather than merely longer or shorter textual outputs. Second, in text-based tasks the cost of reasoning is primarily token count and latency, whereas in closed-loop driving, generated language is an intermediate computation that alters the action trajectory. Unnecessary language can therefore actively degrade driving performance, a harmful effect that BLUE systematically quantifies and that has no direct counterpart in text-based settings. Third, LLM methods modify the language model itself through RL or SFT, while BLUE keeps the VLA backbone frozen and trains only a 0.11M-parameter gate supervised by paired closed-loop driving outcomes rather than task accuracy or token cost.

\section{Additional Experimental Details}
\label{sec:appendix_experimental_details}

This section provides implementation and evaluation details needed for reproducibility.

\subsection{Details of Data Splits}
\label{sec:appendix_data_splits}

We follow the standard data splits and ensure full separation between training and evaluation.

\paragraph{Training Set.}
For all CARLA simulation experiments, data collection for BLUE, including hidden-state extraction, label construction, and gate training, is performed exclusively on routes from the SimLingo training set. For each training route, we run language mode and direct action mode separately through repeated experiments, recording the per-route success rate of each mode. The success rate gap between the two modes determines the training label for that route. We also extract the hidden state at every frame during these runs, which serves as the input feature for gate training. Fail2Drive is used only for direct evaluation and not for training. For NAVSIM, we train on \texttt{navtrain}. The specific route counts, seed configurations, and sampling strategy are detailed in Section~\ref{sec:appendix_implementation_details}.

\paragraph{Evaluation Set.}
We evaluate BLUE on the standard Bench2Drive test split, the full Longest6 v2 benchmark, and Fail2Drive. These CARLA evaluation routes are entirely disjoint from the training routes, and our split is consistent with prior work to ensure fair comparison. No training data, labels, or hidden states are derived from these evaluation routes. For NAVSIM, we evaluate on \texttt{navtest}.

\paragraph{Hyperparameter Selection.}
We do not perform systematic hyperparameter search. All hyperparameters are set with simple heuristics. For example, gate threshold $\theta{=}0.66$ follows from the observation that language impact falls into three natural categories: language-harmful, language-neutral, and language-helpful. These three categories map to three equal intervals on the gate output range $[0, 1]$, so we place the threshold at the boundary of the upper interval. This simple choice already yields strong performance. We provide threshold sensitivity analysis in Section~\ref{sec:analysis}.

\subsection{Details of Benchmarks Considered}
\label{sec:appendix_benchmarks}

We evaluate BLUE on three complementary closed-loop benchmarks built on the CARLA simulator~\cite{dosovitskiy2017carla}. Bench2Drive tests multi-scenario driving over scenario-focused routes, Longest6 v2 evaluates sustained driving quality over longer routes, and Fail2Drive~\cite{gerstenecker2026fail2drive} assesses generalization across 100 standard routes and 100 rare, challenging long-tail routes. Together, they cover scenario diversity, route length, and long-tail generalization. We further evaluate BLUE on NAVSIM~\cite{dauner2024navsim}, which uses real-world driving data, to assess its effectiveness beyond simulation. We provide further details on all four benchmarks below.

\subsubsection{Details of Bench2Drive}
\label{sec:appendix_bench2drive}

Bench2Drive~\cite{jia2024bench2drive} is a closed-loop benchmark built on CARLA v2 for evaluating end-to-end autonomous driving systems across multiple driving skills. It defines 220 evaluation routes distributed across 44 interactive scenario categories, 23 weather conditions, and 12 towns. Each route contains a single safety-critical scenario such as cut-in, overtaking, construction obstacle, or pedestrian crossing. The ego vehicle receives raw sensor inputs and target waypoints, and must drive from the source to the destination within an allotted time without traffic violations.

\paragraph{Multi-Ability Evaluation.} Bench2Drive groups the 44 scenarios into five high-level driving skills: Merging, Overtaking, Emergency Brake, Give Way, and Traffic Sign. Each skill score is the success rate over the corresponding subset of routes, and their average forms the multi-ability mean. This design enables disentangled assessment of individual driving capabilities.

\paragraph{Metrics.} Bench2Drive reports four evaluation metrics. Success Rate and Driving Score capture goal-achieving ability, while Driving Efficiency and Driving Smoothness measure driving quality.

\paragraph{Success Rate (SR).} Success Rate measures the proportion of routes completed within the allotted time without any traffic violation. A route is successful only if the ego vehicle reaches its destination with no recorded infraction. Formally, $\text{SR} = n_{\text{success}} / n_{\text{total}}$, where $n_{\text{success}}$ and $n_{\text{total}}$ denote the number of successful and total routes.

\paragraph{Driving Score (DS).} Driving Score follows the CARLA Leaderboard protocol and combines route completion with infraction penalties:
\begin{equation}
    \text{DS} = \frac{1}{n_{\text{total}}} \sum\nolimits_{i=1}^{n_{\text{total}}} \text{RC}_i \cdot \prod\nolimits_{j} p_{i,j},
\end{equation}
where $\text{RC}_i \in [0, 100]$ is the route completion percentage for route $i$ and $p_{i,j} \in (0, 1]$ is the penalty factor for the $j$-th infraction. Penalties compound multiplicatively, so a single serious infraction can substantially reduce the score.

\paragraph{Driving Efficiency.} Driving Efficiency evaluates whether the ego vehicle maintains a reasonable speed relative to surrounding traffic. At each checkpoint, the ego speed is compared to the average speed of nearby vehicles, yielding a speed percentage. Bench2Drive checks speed every 5\% of the route length across 20 checkpoints to reduce measurement variance. The final metric is the mean speed percentage across all checkpoints: $\overline{v}_\% = (\sum_i v_{\%,i}) / C$, where $C$ is the total number of checkpoints.

\paragraph{Driving Smoothness.} Driving Smoothness evaluates trajectory comfort following the nuPlan protocol~\cite{caesar2021nuplan}. At each frame, six kinematic variables are checked against expert-derived thresholds: longitudinal acceleration, lateral acceleration, yaw rate, yaw acceleration, longitudinal jerk, and jerk magnitude. A frame is smooth only if all six variables fall within their bounds. To mitigate the effect of necessary reactions such as emergency braking, Bench2Drive segments the trajectory into intervals of 20 timesteps and evaluates smoothness per segment. The final score is the ratio of smooth segments to total segments: $\text{Smoothness} = S_{\text{smooth}} / S_{\text{total}}$.

\subsubsection{Details of Longest6 v2}
\label{sec:appendix_longest6}

Longest6 v2~\cite{carla_garage} is a closed-loop benchmark derived from the original Longest6 benchmark~\cite{chitta2022transfuser}. It consists of 36 long routes of 1--2 km each and 7 scenario types. It evaluates whether a driving model can sustain safe and effective performance over longer driving horizons, where early errors and infractions can propagate and compound.
Longest6 v2 reports three standard CARLA Leaderboard metrics.

\paragraph{Driving Score (DS).} Driving Score is defined identically to the Bench2Drive formulation: the average over all routes of the route completion percentage multiplied by the cumulative infraction penalty. Over longer routes, the multiplicative penalty structure makes Driving Score more sensitive to infractions, since more events can accumulate along the route.

\paragraph{Route Completion (RC).} Route Completion measures the average percentage of the prescribed route distance that the ego vehicle successfully traverses before termination. It captures how far the model can drive regardless of infractions.

\paragraph{Infraction Score (IS).} Infraction Score isolates the penalty component by reporting the average product of all per-route infraction multipliers: $\text{IS} = \frac{1}{n_{\text{total}}} \sum_{i} \prod_{j} p_{i,j}$. A higher Infraction Score indicates fewer and less severe violations. Together with Route Completion, Infraction Score enables diagnosis of whether low Driving Score stems from incomplete routes or from frequent infractions.

\subsubsection{Details of Fail2Drive}
\label{sec:appendix_fail2drive}

Fail2Drive~\cite{gerstenecker2026fail2drive} is a closed-loop benchmark built on CARLA 0.9.15 for evaluating generalization under controlled distribution shifts. It contains 200 short routes in Town 13, with an average length of 219 meters. The routes form 100 pairs, each consisting of an in-distribution route with a standard CARLA scenario and a generalization route with a targeted long-tail shift. Within each pair, the road geometry, spawn points, traffic conditions, and driving objective remain fixed, allowing the performance difference to isolate the effect of the distribution shift.

\paragraph{Generalization Scenarios.} Fail2Drive introduces 17 scenario classes organized into four categories: robustness, visual generalization for lateral control, visual generalization for longitudinal control, and behavioral generalization. These categories test whether driving models remain reliable under irrelevant visual changes, unseen obstacle appearances and layouts, novel causal objects such as crossing animals, and unfamiliar behaviors such as stopping for a fully blocked road.

\paragraph{Metrics.} Fail2Drive follows the Bench2Drive protocol and reports Driving Score and Success Rate on both the in-distribution and generalization routes. Because its routes are short and usually finishable, RC is not reported as a separate metric. Fail2Drive additionally summarizes the two primary metrics using their harmonic mean:
\begin{equation}
    \text{HM} = \frac{2 \cdot \text{DS} \cdot \text{SR}}{\text{DS} + \text{SR}}.
\end{equation}
For each metric $M$, the relative generalization change is computed as $100(M_{\text{gen}}-M_{\text{ID}})/M_{\text{ID}}$. This paired comparison directly measures how much performance changes when the model encounters a controlled long-tail shift.

\subsubsection{Details of NAVSIM}
\label{sec:appendix_navsim}

NAVSIM~\cite{dauner2024navsim} is a data-driven, non-reactive benchmark for evaluating autonomous driving planners on real-world sensor data. It is built on OpenScene, a redistribution of the nuPlan dataset~\cite{caesar2021nuplan}, and provides standardized \texttt{navtrain} and \texttt{navtest} splits with approximately 103K and 12K samples, respectively. Each sample may include multi-view camera images, LiDAR, ego motion states, and a navigation goal.

\paragraph{Non-Reactive Simulation.} Given the initial sensor observation, an agent predicts a trajectory over a 4-second horizon and is not queried again during the rollout. An LQR controller and a kinematic bicycle model execute the predicted trajectory at 10 Hz, while surrounding agents follow their recorded trajectories. This design enables scalable and deterministic evaluation on real-world data, while simulation-based outcomes provide a closer approximation to closed-loop driving quality than displacement errors alone.

\paragraph{Predictive Driver Model Score (PDMS).} NAVSIM evaluates safety, progress, and comfort through five normalized subscores: No At-Fault Collision (NC), Drivable Area Compliance (DAC), Time to Collision (TTC), Ego Progress (EP), and Comfort (C). NC and DAC serve as multiplicative penalties, while TTC, EP, and C are combined using a weighted average:
\begin{equation}
    \text{PDMS} = \text{NC} \cdot \text{DAC} \cdot
    \frac{5\text{EP} + 5\text{TTC} + 2\text{C}}{12}.
\end{equation}
NC penalizes at-fault collisions, DAC measures whether the trajectory remains within the drivable area, TTC captures near-term collision risk, EP measures progress relative to a safe reference planner, and C evaluates acceleration and jerk. We also report Driving Direction Compliance as an auxiliary diagnostic metric. NAVSIM complements the three CARLA benchmarks by testing BLUE on recorded real-world observations, although its non-reactive evaluation does not model long-term interaction or compounding closed-loop errors.

\subsection{Details of Baselines Considered}
\label{sec:appendix_baselines}
We compare BLUE against a wide range of published methods spanning end-to-end driving and vision-language-action approaches. Table~\ref{tab:driving_results} lists the sensor configuration and training labels for each method. As shown in the table, most baselines rely on surround cameras, LiDAR, or dense auxiliary labels such as 3D object detection, HD maps, semantic segmentation, and depth. In contrast, BLUE uses only front-view camera without LiDAR and requires only language annotations, yet achieves the best closed-loop results on two benchmarks. Below we briefly describe each baseline.

\paragraph{UniAD-Base \cite{hu2023planning}} unifies perception, prediction, and planning into a single network, where all tasks communicate through unified query interfaces and are jointly optimized toward planning. It uses six surround cameras and trains with 3D object detection, map, and segmentation labels.

\paragraph{TF++ \cite{zimmerlin2024hidden}} analyzes training data biases and proposes a label-change-based frame selection criterion to compress datasets without losing important information. It uses a front-view camera with LiDAR and trains with object detection, map, segmentation, and depth labels.

\paragraph{MomAD \cite{song2025don}} introduces trajectory momentum and perception momentum to stabilize long-horizon planning by selecting planning queries topologically consistent with historical paths and fusing them with historical context. It uses six cameras with object detection and map labels.

\paragraph{DriveTransformer \cite{jia2025drivetransformer}} replaces the sequential perception-prediction-planning pipeline with parallel task interaction, where agent, map, and planning queries directly attend to each other at every block. It uses six cameras with object detection and map labels.

\paragraph{Hydra-NeXt \cite{li2025hydra}} adopts a multi-branch framework that unifies trajectory prediction, control prediction, and trajectory refinement to bridge the gap between open-loop training and closed-loop driving. It uses two cameras without auxiliary perception labels.

\paragraph{Raw2Drive \cite{yang2026raw2drive}} follows a dual-stream model-based reinforcement learning approach, first training a privileged world model and then aligning a raw-sensor world model to it through a guidance mechanism. It uses six cameras with map and segmentation labels.

\paragraph{DiffusionDrive \cite{liao2025diffusiondrive}} applies a truncated diffusion policy with multi-mode anchors, compressing the denoising process to produce diverse driving actions in only a few steps. It uses multiple cameras with LiDAR and trains with object and segmentation labels.

\paragraph{ORION \cite{fu2025orion}} bridges semantic reasoning and numerical trajectory output by combining a QT-Former for long-term context aggregation, a large language model for scene reasoning, and a generative planner for trajectory prediction. It uses six cameras with object detection, map, and language labels.

\paragraph{AutoVLA \cite{zhou2026autovla}} unifies reasoning and action generation within a single autoregressive model by tokenizing continuous trajectories into discrete action tokens. It supports fast and slow thinking modes and uses GRPO-based reinforcement fine-tuning to reduce unnecessary reasoning. It uses a front-view camera with language labels.

\paragraph{SimLingo \cite{renz2025simlingo}} is a camera-only vision-language model that jointly handles closed-loop driving, vision-language understanding, and language-action alignment. It uses a single front-view camera with language annotations and serves as the primary backbone for BLUE.

\paragraph{HiP-AD \cite{tang2025hip}} introduces multi-granularity planning queries that integrate spatial, temporal, and driving-style waypoints, and uses deformable attention to retrieve image features based on physical trajectory locations. It uses six cameras with object detection, map, and depth labels.

\paragraph{ReCogDrive \cite{li2025recogdrive}} combines an autoregressive cognitive model with a diffusion planner, where the former provides driving reasoning priors and the latter generates continuous trajectories. It uses six cameras with language labels.

\paragraph{GeRo \cite{yasarla2026generative}} extends VLA models with language-conditioned autoregressive generation of future traffic scenes, encoding ego and agent dynamics into shared latent tokens and stabilizing long-horizon rollouts with a consistency loss. It uses six cameras with object detection, map, and language labels.

\paragraph{DeLL \cite{du2026deconfounded}} addresses lifelong learning in end-to-end driving through a Dirichlet process mixture model that builds dynamic knowledge spaces, combined with front-door causal adjustment to suppress spurious correlations. It uses a front-view camera with LiDAR and trains with object detection and segmentation labels.

\paragraph{R2SE \cite{liu2026reinforced}} proposes a three-stage pipeline: generalist pretraining with hard-case identification, residual reinforcement fine-tuning on difficult scenarios, and self-aware adapter expansion that routes between generalist and specialist policies at test time. It uses a front-view camera with LiDAR and trains with object detection, map, segmentation, and depth labels.

\paragraph{AutoMoT \cite{huang2026automot}} unifies reasoning and action generation within a mixture-of-transformers architecture, where a frozen understanding expert and a high-frequency action expert share a joint attention space for asynchronous fast-slow inference. It uses a front-view camera with LiDAR.

\paragraph{BevAD \cite{holtz2026matters}} re-examines common architectural patterns for closed-loop driving and identifies effective combinations of spatial bottleneck compression, decoupled trajectory representation, and diffusion-based planning. It uses six cameras with object detection labels.

\paragraph{CriticVLA \cite{yang2026judge}} extends VLA models from acting to judging: it generates a rough trajectory and then uses the VLA as a critic to evaluate and refine the plan through single-step optimization. It uses a single front-view camera with language labels and serves as a secondary backbone for validating BLUE.

\paragraph{TakeVLA \cite{gao2026learning}} improves VLA driving through post-training on expert takeover data, shifting language supervision to the period before takeover moments so that the model learns to anticipate hazards early. It uses a single front-view camera with language labels.

\subsection{Details of Label Construction}
\label{sec:appendix_labeling_details}

The gate requires binary labels indicating whether each frame benefits from language generation. A key advantage of our labeling approach is that it requires no human annotation: all labels are derived automatically from closed-loop evaluation outcomes by comparing the two modes. We construct labels at two granularities, route-level and frame-level, and apply temporal redundancy cleaning before training.

\subsubsection{Route-Level Labels}
\label{sec:appendix_route_labels}

We run language mode and direct action mode on each training route with $|\mathcal{S}|{=}5$ random seeds. The cross-seed success rate for mode $m$ on route $r$ is $\overline{\mathrm{SR}}_m^{(r)} = \frac{1}{|\mathcal{S}|}\sum_{s \in \mathcal{S}} \mathrm{SR}_m^{(r,s)}$. As defined in Eq.~\ref{eq:label_route}, a route receives label $y_r{=}1$ when the language advantage exceeds a margin threshold:
\begin{equation}\label{eq:language_advantage}
    \Delta\overline{\mathrm{SR}}_r = \overline{\mathrm{SR}}_{\text{lang}}^{(r)} - \overline{\mathrm{SR}}_{\text{direct}}^{(r)} > \tau,
\end{equation}
where $\tau{=}10\%$ ensures that language mode is activated only when it provides a sufficiently large performance gain. Routes that do not meet this condition are labeled $y_r{=}0$, defaulting to the faster direct action mode. Under route-level labeling, all frames within a route share the same label $y_r$.

\subsubsection{Frame-Level Labels}
\label{sec:appendix_frame_labels}

Route-level labels assign a uniform label to all frames in a route, but in practice, even on a language-beneficial route, only certain segments truly require language. To provide finer supervision, we identify critical regions $\mathcal{C}_r$ where the two modes exhibit the largest behavioral divergence. The core idea is straightforward: we spatially compare how the vehicle behaves under the two modes and mark the locations where their behaviors differ the most.
Concretely, for each language-beneficial route, we collect 2D ego-vehicle trajectories from all seeds under both modes and overlay them on a uniform spatial grid. For each grid cell $\mathbf{g}$, we measure the normalized cross-mode behavioral difference for each signal channel $k$:
\begin{equation}\label{eq:signal_diff}
    \Delta_k(\mathbf{g}) = \frac{|\bar{v}_{k}^{\text{lang}}(\mathbf{g}) - \bar{v}_{k}(\mathbf{g})|}{\max_{\mathbf{g}'} |\bar{v}_{k}^{\text{lang}}(\mathbf{g}') - \bar{v}_{k}(\mathbf{g}')| + \epsilon},
\end{equation}
where $\bar{v}_{k}(\mathbf{g})$ and $\bar{v}_{k}^{\text{lang}}(\mathbf{g})$ are the seed-averaged behavioral signals at cell $\mathbf{g}$ under direct action mode and language mode, respectively. The behavioral channels include speed, acceleration, heading, and trajectory spread. Additionally, we construct an infraction channel by placing spatial kernels at infraction locations from simulation logs, measuring where the two modes differ in safety outcomes.

We aggregate all channels into a per-cell criticality score $c(\mathbf{g}) = \sum_{k} w_k \cdot \Delta_k(\mathbf{g})$, threshold the resulting map, and retain the top spatial regions as critical regions $\mathcal{C}_r$. A frame receives $y_{r,t}{=}1$ only if both $y_r{=}1$ and its spatial position falls within $\mathcal{C}_r$ (Eq.~\ref{eq:label_frame}). During training, we mix route-level and frame-level samples so that the gate learns both coarse route preference and fine-grained activation patterns. We will fully release the labeling code for complete implementation details.

\subsubsection{Temporal Redundancy Cleaning}
\label{sec:appendix_temporal_cleaning}

When the vehicle remains still, such as while waiting at a red light, many consecutive frames contain almost the same scene and therefore produce almost identical hidden states. If we keep all of them, these low-motion moments would be overrepresented during training. We detect such redundant segments by computing cosine similarity between adjacent hidden-state vectors:
\begin{equation}\label{eq:cosine_redundancy}
    \mathrm{sim}(\mathbf{h}_t, \mathbf{h}_{t+1}) = \frac{\mathbf{h}_t^\top \mathbf{h}_{t+1}}{\|\mathbf{h}_t\| \cdot \|\mathbf{h}_{t+1}\|} \geq 0.99.
\end{equation}
Adjacent frames whose similarity exceeds this threshold are treated as one redundant segment of length $L$. From each segment, we keep only $k = \max(2,\; \lceil L^{\alpha} \rceil)$ evenly spaced representative samples, where $\alpha{=}0.5$. This sublinear schedule ensures short segments retain at least two samples while preventing long idle periods from overwhelming the dataset. After cleaning, the training set reduces by approximately 15\% in frame count while preserving coverage of all driving states.

\subsection{Implementation and Hyperparameters}
\label{sec:appendix_implementation_details}

This subsection reports all implementation details needed to reproduce the gate training pipeline.

\subsubsection{Data Collection}
\label{sec:appendix_hidden_state_extraction}

We extract the hidden state $\mathbf{h} \in \mathbb{R}^d$ from the last transformer layer at the final token position of a prompt-only forward pass, where $d{=}896$ is the hidden dimension of InternVL2-1B~\cite{chen2024internvl}. For both language mode and direct action mode, we forward only the prompt tokens, including visual tokens and system instructions, and collect$\mathbf{h}$ from the same position. This ensures that hidden states from the two modes are aligned and that the gate receives comparable inputs regardless of which mode produced them.
All data are collected on approximately 400 routes sampled from the SimLingo training set, stratified by scenario difficulty to cover both common and rare driving situations. For each route, we run both modes separately with 5 random seeds, recording per-frame hidden states and per-route success outcomes. The success rate gap determines training labels as described in Section~\ref{sec:appendix_labeling_details}.
BLUE does not require a dedicated data collection effort. In autonomous driving development, validating a model already involves running closed-loop evaluations under different configurations. BLUE simply reuses hidden states and outcomes from these routine runs, making data collection a natural byproduct of the standard pipeline rather than an additional cost.

\subsubsection{Gate Architecture and Training}
\label{sec:appendix_gate_training}
The gate is a single-hidden-layer MLP that maps $\mathbf{h}$ to a scalar activation probability:
\begin{equation}
    \mathbf{z} = W_1 \mathbf{h} + b_1, \quad p(\mathbf{h}) = \sigma(W_2 \,\tilde{\mathbf{z}} + b_2),
\end{equation}
where $\tilde{\mathbf{z}}$ denotes $\mathbf{z}$ after ReLU activation and dropout, $W_1 {\in} \mathbb{R}^{m \times d}$, $W_2 {\in} \mathbb{R}^{1 \times m}$, and $\sigma$ is the sigmoid function. We set hidden dimension $m{=}128$ and dropout rate 0.5, yielding 0.11M total trainable parameters.
We optimize with Adam using learning rate 0.001, weight decay 0.05, cosine annealing schedule, and batch size 512 for 100 epochs. 
The training set contains approximately 670K frames.

\subsubsection{Computational Cost}
\label{sec:appendix_computational_cost}
The total overhead of BLUE is minimal, largely because its data collection naturally aligns with the standard autonomous driving development workflow. Models must undergo closed-loop road testing before deployment, during which driving outcomes under different configurations are routinely recorded. BLUE simply records hidden states alongside these existing evaluation runs and derives labels automatically by comparing route success rates between the two modes. This means data collection requires minimal additional effort beyond routine validation, no reward engineering, and no human labeling. The raw GPU time consumed by these evaluation runs is approximately 1200 A100 GPU hours, but the vast majority is not a net addition to the development budget. Gate training requires less than 0.1 GPU hours on a single GPU, and the inference overhead is negligible as it adds only a single MLP forward pass per frame.
Separately, the language impact analysis in Appendix~\ref{sec:appendix_extra_settings} consumes approximately $\sim$2000 A100 GPU hours in total, covering repeated closed-loop experiments across multiple configurations.

\subsection{Details of Baseline Gates}
\label{sec:appendix_baseline_gates}

This section describes the construction of alternative gating strategies evaluated in Table~\ref{tab:gate_baselines}. All baseline gates share the same inference procedure as BLUE: at each frame, the gate produces a binary decision that determines whether to activate language generation or predict actions directly.

\paragraph{Kinematic Gates.}
We design three rule-based gates, each using a single kinematic feature available at inference time: vehicle speed in m/s, acceleration magnitude as the L2 norm of the 3D acceleration vector in m/s\textsuperscript{2}, and steering angle as the absolute value of the previous control output normalized to 0--1. A gate activates language generation whenever the corresponding feature exceeds a predefined threshold, selected so that the language activation ratio approximates that of BLUE.

\paragraph{Complexity-based Gate.}
Rather than relying on frame-level kinematic signals, the complexity-based gate uses route-level scenario complexity as supervision to train a gate. We first compute a composite complexity score for each training route based on structured features extracted from the CARLA scenario configuration file:
\begin{equation}
s = w_1 \cdot f_{\text{scen}} + w_2 \cdot f_{\text{wea}} + w_3 \cdot f_{\text{flow}} + w_4 \cdot f_{\text{freq}},
\end{equation}
where $f_{\text{scen}}$ reflects the number of sub-scenarios in the route normalized to [0, 1], $f_{\text{wea}}$ aggregates fog density, precipitation intensity, and a night-driving indicator, $f_{\text{flow}}$ indicates the presence of dynamic traffic flows requiring gap-finding maneuvers, and $f_{\text{freq}}$ captures periodic opposing vehicle interactions. We set $w_1{=}0.30$, $w_2{=}0.30$, $w_3{=}0.25$, $w_4{=}0.15$. Routes with $s \geq \tau$ are labeled as complex and the rest as simple. All frames within a route share the same label. We then train a gate with the same MLP architecture as BLUE, supervised with these complexity-derived labels. At inference, this gate predicts whether the current frame corresponds to a complex scenario and activates language generation accordingly.

\paragraph{Random Gate.}
The random gate activates language generation for each frame with a fixed probability, matched to the language activation ratio of BLUE. This baseline isolates the effect of selectively choosing when to generate language from the effect of simply reducing language frequency.

\begin{table*}[t!]
\centering
\renewcommand{\arraystretch}{1.08}
\resizebox{\textwidth}{!}{%
\begin{tabular}{@{}llcclc cc@{}}
\toprule
\multirow{2}{*}{\textbf{Method}} & \multicolumn{5}{c}{\textbf{Details}} & \multicolumn{2}{c}{\textbf{Metrics}} \\ \cmidrule(lr){2-6} \cmidrule(lr){7-8}
& Expert & Camera & LiDAR & Labels & T-Param. & SR (\%) $\uparrow$ & DS $\uparrow$ \\ \midrule
TCP* \cite{wu2022trajectory} & Think2Drive & 3$\times$ & \xmark & - & $\approx$\,26\,M & 15.00 & 40.70 \\
TCP-traj* \cite{wu2022trajectory} & Think2Drive & 3$\times$ & \xmark & - & $\approx$\,26\,M & 30.00 & 59.90 \\
VAD \cite{jiang2023vad} & Think2Drive & 6$\times$ & \xmark & O,M & $\geq$\,25\,M & 15.00 & 42.35 \\
UniAD-Base \cite{hu2023planning} & Think2Drive & 6$\times$ & \xmark & O,M,S & $\geq$\,59\,M & 16.36 & 45.81 \\
ThinkTwice* \cite{jia2023think} & Think2Drive & 6$\times$ & \xmark & S,D & $\approx$\,120\,M & 31.23 & 62.44 \\
DriveAdaptor* \cite{jia2023driveadapter} & Think2Drive & 6$\times$ & \xmark & M,S,D & $\approx$\,135\,M & 33.08 & 64.22 \\
GenAD \cite{zheng2024genad} & Think2Drive & 6$\times$ & \xmark & O,M & $\geq$\,25\,M & 15.90 & 44.81 \\
TF++ \cite{zimmerlin2024hidden} & PDM-Lite & 1$\times$ & \cmark & O,M,S,D & $\geq$\,39\,M & 67.27 & 84.21 \\
MomAD \cite{song2025don} & Think2Drive & 6$\times$ & \xmark & O,M & $\geq$\,25\,M & 18.11 & 47.91 \\
DriveTrans \cite{jia2025drivetransformer} & Think2Drive & 6$\times$ & \xmark & O,M & $\approx$\,646\,M & 35.01 & 63.46 \\
ETA \cite{hamdan2025eta} & Think2Drive & 1$\times$ & \xmark & - & $\geq$\,300\,M & 48.33 & 74.33 \\
Hydra-NeXt \cite{li2025hydra} & Think2Drive & 2$\times$ & \xmark & - & $\geq$\,25\,M & 50.00 & 73.86 \\
Raw2Drive \cite{yang2026raw2drive} & - & 6$\times$ & \xmark & M,S & $\geq$\,25\,M & 50.24 & 71.36 \\
DiffusionDrive \cite{liao2025diffusiondrive} & - & 3$\times$ & \cmark & O,S & $\approx$\,60\,M & 52.72 & 77.68 \\
ORION \cite{fu2025orion} & Think2Drive & 6$\times$ & \xmark & O,M,L & $\geq$\,300\,M & 54.62 & 77.74 \\
AutoVLA \cite{zhou2026autovla} & PDM-Lite & 1$\times$ & \xmark & L & $\geq$\,1.5\,B & 57.73 & 78.84 \\
SimLingo \cite{renz2025simlingo} & PDM-Lite & 1$\times$ & \xmark & L & $\geq$\,300\,M & 67.27 & 85.07 \\
HiP-AD \cite{tang2025hip} & Think2Drive & 6$\times$ & \xmark & O,M,D & $\approx$\,97\,M & 69.09 & 86.77 \\
ReCogDrive \cite{li2025recogdrive} & Think2Drive & 6$\times$ & \xmark & L & $\geq$\,2\,B & 45.45 & 71.36 \\
GeRo \cite{yasarla2026generative} & Think2Drive & 6$\times$ & \xmark & O,M,L & $\geq$\,3\,B & 60.10 & 81.90 \\
DeLL \cite{du2026deconfounded} & PDM-Lite & 1$\times$ & \cmark & O,S & $\geq$\,38\,M & 68.63 & 86.86 \\
R2SE \cite{liu2026reinforced} & PDM-Lite & 1$\times$ & \cmark & O,M,S,D & $\geq$\,39\,M & 69.54 & 86.28 \\
AutoMoT \cite{huang2026automot} & PDM-Lite & 1$\times$ & \cmark & - & $\approx$\,1.6\,B & 70.00 & 87.34 \\
BevAD \cite{holtz2026matters} & PDM-Lite & 6$\times$ & \xmark & O & $\geq$\,25\,M & 72.73 & 88.11 \\
CriticVLA \cite{yang2026judge} & PDM-Lite & 1$\times$ & \xmark & L & $\geq$\,300\,M & 73.33 & 88.02 \\
TakeVLA \cite{gao2026learning} & PDM-Lite & 1$\times$ & \xmark & L & $\geq$\,300\,M & 73.73 & 89.72 \\
\addlinespace[2pt]
\midrule
\rowcolor{bbb!15}\textbf{BLUE (SimLingo)} & PDM-Lite & 1$\times$ & \xmark & L & \textbf{0.11\,M} & \textbf{76.18{\scriptsize$\pm$0.64}} & \textbf{90.58{\scriptsize$\pm$0.12}} \\
$\Delta$ vs. SimLingo & - & - & - & - & - & +8.91 & +5.51 \\
\addlinespace[2pt]
\rowcolor{bbb!15}\textbf{BLUE (CriticVLA)} & PDM-Lite & 1$\times$ & \xmark & L & \textbf{0.11\,M} & 76.04{\scriptsize$\pm$0.38} & 90.37{\scriptsize$\pm$0.14} \\
$\Delta$ vs. CriticVLA & - & - & - & - & - & +2.71 & +2.35 \\ \bottomrule
\end{tabular}%
}
\caption{Full results on Bench2Drive, showing the complete comparison between BLUE and 26 baselines. BLUE (SimLingo) ranks first and BLUE (CriticVLA) ranks second in terms of closed-loop success rate (SR) and driving score (DS). T-Param.\ reports driving-task trainable parameters; we use published values ($\approx$) where available and conservative lower bounds ($\geq$) derived from the minimum size of trained components. Both BLUE variants train only a 0.11\,M gate while keeping the entire VLA backbone frozen, yet surpass methods that employ multi-camera setups, LiDAR, or dense auxiliary labels (O: 3D object detection, M: map, S: semantic segmentation, D: depth, L: language), using only a single front-view camera with language annotations.}
\label{tab:driving_results_full}
\vspace{-5pt}
\end{table*}

\section{Additional Experimental Results}
\label{sec:appendix_additional_results}

The main text reports key results on Bench2Drive. This section provides the complete comparison tables, covering 26 baselines alongside our BLUE on both SimLingo and CriticVLA backbones.

\subsection{Full Bench2Drive Comparison}
\label{sec:appendix_full_b2d}

Table~\ref{tab:driving_results_full} presents the full Bench2Drive comparison across 26 methods. For completeness, we include both BLUE (SimLingo) and BLUE (CriticVLA) in a single table. Both variants use only a single front-view camera and language annotations, yet outperform their respective backbones by clear margins. BLUE (SimLingo) achieves the overall best SR and DS among all methods, while BLUE (CriticVLA) also surpasses its backbone and ranks among the top entries. This confirms that the on-demand language use strategy is effective across different VLA architectures under the same evaluation protocol.

\begin{table*}[t]
\centering
\small
\renewcommand{\arraystretch}{1.1}
\begin{tabular}{@{}lllccc@{}}
\toprule
\textbf{Category} & \textbf{Setting} & \textbf{Method} & \textbf{Helpful (\%)} & \textbf{Neutral (\%)} & \textbf{Harmful (\%)} \\
\midrule
\multirow{2}{*}{Original} & \multirow{2}{*}{SimLingo} & Threshold & 14.5 & 61.8 & 23.6 \\
 & & Sign test & 14.5 & 62.7 & 22.7 \\
\midrule
\multirow{2}{*}{Granularity} & \multirow{2}{*}{Brief} & Threshold & 21.8 & 52.7 & 25.5 \\
 & & Sign test & 21.8 & 52.7 & 25.5 \\
\midrule
\multirow{2}{*}{Language} & \multirow{2}{*}{Chinese} & Threshold & 18.6 & 58.2 & 23.2 \\
 & & Sign test & 15.9 & 67.7 & 16.4 \\
\midrule
\multirow{2}{*}{Model} & \multirow{2}{*}{CriticVLA} & Threshold & 19.5 & 52.3 & 28.2 \\
 & & Sign test & 18.2 & 55.5 & 26.4 \\
\bottomrule
\end{tabular}
\caption{Language impact categorization across four experimental settings, each evaluated with both the threshold method and the sign test. The two methods yield closely aligned results. Across all settings, the majority of routes are neutral, and language-harmful routes consistently equal or outnumber language-helpful routes.}
\label{tab:language_impact_full}
\end{table*}

\subsection{Language Impact Analysis}
\label{sec:appendix_extra_settings}

This section details the experimental settings and statistical methods used to categorize each route as language-helpful, language-neutral, or language-harmful, as summarized in Table~\ref{tab:annotation} of the main text. All analyses in this section are conducted on the Bench2Drive evaluation set. The gate training uses only training routes with no overlap with the evaluation set, as detailed in Appendix~\ref{sec:appendix_data_splits}.

\subsubsection{Experimental Settings}

We evaluate the language impact under four settings that vary annotation granularity, annotation language, and VLA backbone. All settings are evaluated on the full 220 routes of Bench2Drive \cite{jia2024bench2drive}, running each route with and without language generation across multiple random seeds. We classify each route using two complementary methods: a threshold method based on effect size and a binomial sign test for statistical significance. As Table~\ref{tab:language_impact_full} shows, both methods produce consistent categorizations across all four settings.

\paragraph{SimLingo.} The default setting uses the official SimLingo \cite{renz2025simlingo} checkpoint trained with English annotations at normal granularity. The annotations describe surrounding objects, traffic conditions, and intended maneuvers at each frame.

\paragraph{Brief.} To test whether annotation granularity affects the language impact distribution, we retrain SimLingo with brief English annotations that retain only action-relevant instructions and remove detailed scene descriptions. The model architecture, training procedure, and evaluation protocol are identical to the default SimLingo setting.

\paragraph{Chinese.} To further test whether the pattern holds under additional annotation settings, we retrain SimLingo with Chinese annotations translated from the default English version. The translated annotations preserve the same content and structure. All other training details remain the same.

\paragraph{CriticVLA.} To test whether the pattern generalizes across VLA architectures, we replace the backbone with CriticVLA \cite{yang2026judge}. Unlike SimLingo, which generates language reasoning before predicting actions in a single pass, CriticVLA first produces an initial trajectory and then uses language-based critique to refine the plan.

\subsubsection{Statistical Methods}

We apply two complementary methods to classify each route. The first provides an intuitive effect-size criterion, and the second offers formal statistical validation.

\paragraph{Threshold method.} For each route $r$, we compute the cross-seed average success rate under language mode $\overline{\mathrm{SR}}_{\text{lang}}^{(r)}$ and under direct action mode $\overline{\mathrm{SR}}_{\text{direct}}^{(r)}$, and take their difference $\Delta^{(r)} = \overline{\mathrm{SR}}_{\text{lang}}^{(r)} - \overline{\mathrm{SR}}_{\text{direct}}^{(r)}$. A route is classified as language-helpful if $\Delta > \tau$, language-harmful if $\Delta < -\tau$, and language-neutral otherwise, where $\tau{=}0.1$ requires the success rate gap to exceed 10\% for a route to be considered meaningfully affected.

\paragraph{Sign test.} For each route $r$, we construct all cross-seed pairings between language seeds and direct action seeds, yielding $n_{\text{lang}} \times n_{\text{direct}}$ pairs per route. Among these, we count the discordant pairs: $n_+$ pairs where language succeeds and direct fails, and $n_-$ pairs where language fails and direct succeeds. Concordant pairs are excluded. Under the null hypothesis that the two modes are equivalent, $n_+$ follows a Binomial$(n_+ + n_-, 0.5)$ distribution. We apply a one-sided binomial exact test at significance level $\alpha{=}0.15$: a route is classified as language-helpful if $P(X \geq n_+) < \alpha$, language-harmful if $P(X \geq n_-) < \alpha$, and language-neutral otherwise. Routes with zero discordant pairs are classified as neutral. We do not apply multiple testing correction because our conclusion depends on the aggregate proportion of the three categories across all routes, not on the classification of any single route.

\subsubsection{Detailed Results}

Table~\ref{tab:language_impact_full} reports the classification results under both methods across all four settings. The two methods produce highly consistent categorizations. In all settings, the majority of routes fall into the neutral category, and the number of language-harmful routes consistently equals or exceeds the number of language-helpful routes. This pattern holds regardless of annotation granularity, annotation language, or VLA backbone, confirming that the complementarity between the two modes is a general property rather than an artifact of a specific configuration.
The agreement between the two methods confirms that our categorization is robust to the choice of statistical criterion, and that selective language activation is warranted across diverse configurations.

\paragraph{Per-scenario results.} Beyond aggregate route counts, Figure~\ref{fig:scenario_language_effect_all_settings} shows how language impact is distributed across scenario categories under the four settings. The scenario-level view reveals that helpful and harmful effects are not confined to a single scenario family. Many categories are dominated by neutral routes, while language-sensitive cases appear as scattered helpful and harmful groups whose locations can change with the setting. Table~\ref{tab:scenario_index_mapping} maps each scenario index in the figure to its Bench2Drive scenario name. These results support the main conclusion that language generation should be selected on demand rather than used by default at every frame.

\subsubsection{Discussion}
The cross-setting results above reveal two practical implications for gate design: why each backbone requires its own gate, and why scenario complexity alone cannot replace hidden-state conditioning.

\begin{figure*}[t]
  \centering
  \includegraphics[width=0.98\textwidth]{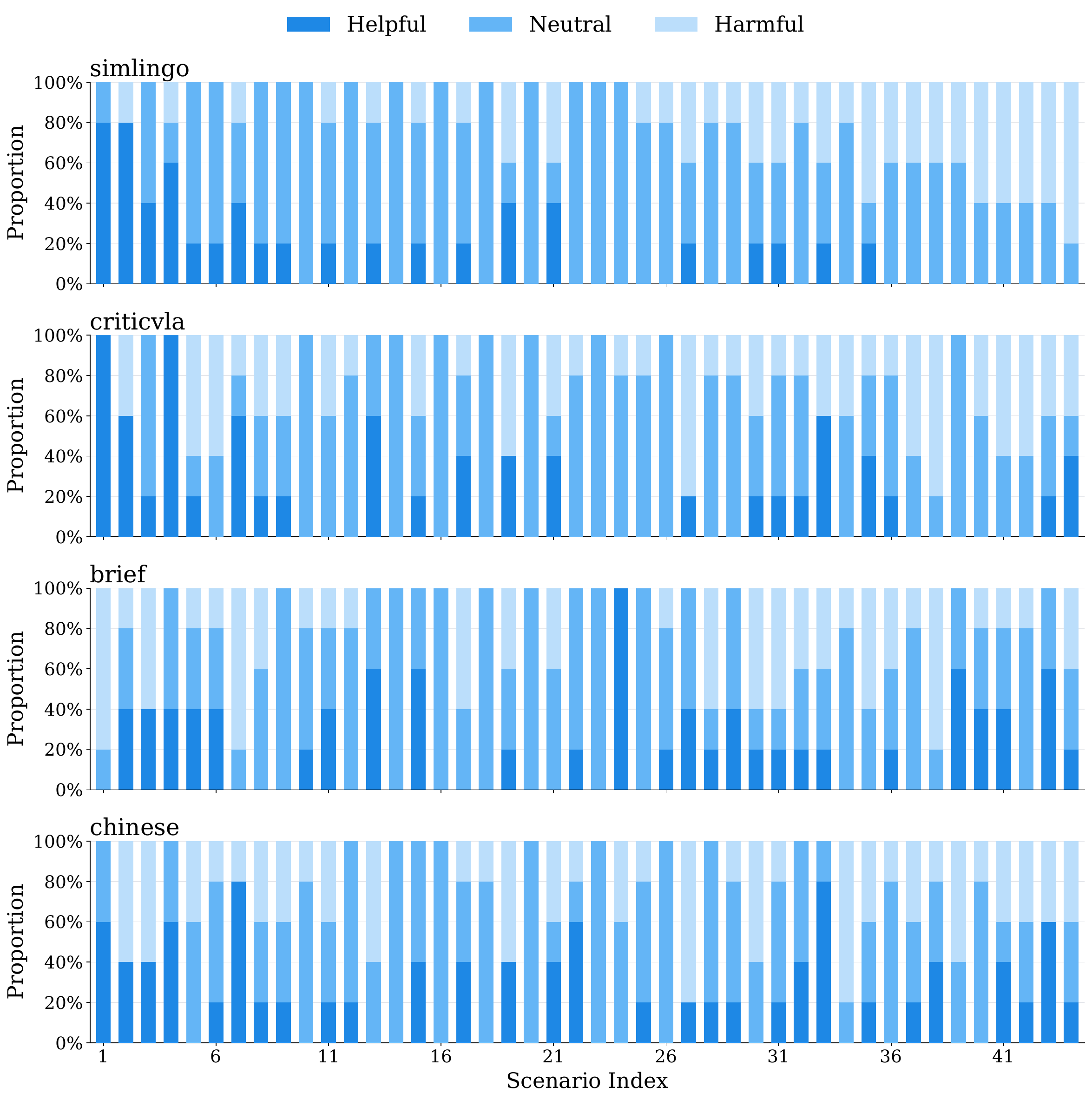}
  \vspace{-6pt}
  \caption{Per-scenario distribution of language effects under four settings. Each bar denotes one Bench2Drive scenario category; all four panels share the same scenario order for direct comparison. Colors indicate routes where language generation is helpful, neutral, or harmful. The scenario index mapping is provided in Table~\ref{tab:scenario_index_mapping}.}
  \label{fig:scenario_language_effect_all_settings}
  \vspace{-5pt}
\end{figure*}

\paragraph{Why each backbone needs its own gate.} Figure~\ref{fig:scenario_language_effect_all_settings} shows that the helpful, neutral, and harmful distributions change across settings, with a clear shift when moving from SimLingo to CriticVLA. If language utility were determined only by the route, the same scenario index would show similar patterns across backbones. Instead, the pattern changes, indicating that language utility also depends on how each backbone uses language and represents the scene before acting. A gate trained for one backbone therefore learns a readout tied to that model, rather than a universal rule. BLUE trains a separate gate for each backbone, but this adds little cost. As discussed in Section~\ref{sec:limitations}, the backbone remains frozen, the gate is a lightweight MLP, and labels come from routine evaluations.

\paragraph{Why scenario complexity is insufficient.} Figure~\ref{fig:scenario_language_effect_all_settings} also explains why a gate based only on scenario complexity cannot work reliably. All four settings share the same set of driving scenarios, yet where language helps or hurts shifts when the model or annotation changes. This means the driving data alone, including route type and scenario complexity, cannot determine whether language will help. Recent adaptive VLA methods, including AutoVLA and AdaThinkDrive \cite{zhou2026autovla, luo2025adathinkdrive}, often reduce reasoning in straightforward scenes and keep more reasoning in difficult ones. Our results show that this heuristic misses the key signal: language generation may help on a given scenario under one model but hurt under another. Complexity alone is insufficient to predict when language will help. BLUE therefore conditions the decision on model hidden states rather than on scenario complexity alone.

\section{Additional Analysis}
\label{sec:appendix_language_content}

This section provides additional analysis on the language content generated by SimLingo.

\subsection{Language Content Visualization}
\label{sec:appendix_language_content_summary}

\begin{figure*}[p]
  \centering
  \includegraphics[width=0.96\textwidth]{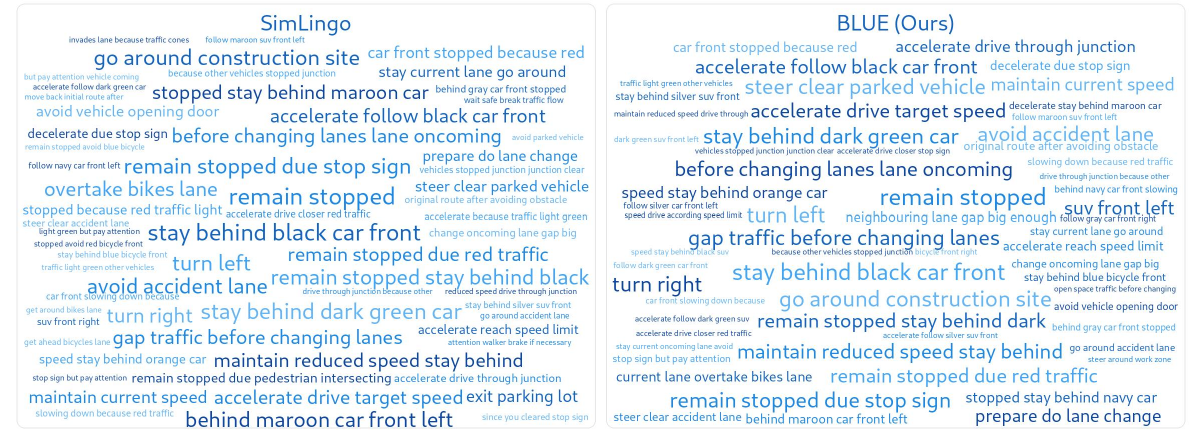}
  \vspace{-5pt}
  \caption{N-gram phrase clouds of generated language during closed-loop driving. Left: SimLingo generates language at every frame. Right: BLUE generates language only when the gate activates. Both panels show the most frequent meaningful phrases (2--5 words) extracted from all evaluation routes.}
  \label{fig:wordcloud}
\end{figure*}

We visualize the language content generated during closed-loop driving through word clouds, providing readers with an intuitive impression of the vocabulary produced by language generation. Figure \ref{fig:wordcloud} presents the word clouds of generated language, where the left panel corresponds to the original SimLingo and the right panel corresponds to BLUE with language generated when the gate activates.

\section{Additional Statements}
\label{sec:appendix_impact_disclosure}

We provide statements on artifact licensing, potential risks, broader impact, and the use of large language models in preparing this work.

\paragraph{Licenses and Terms of Use}
\label{sec:appendix_license}
We evaluate BLUE on four benchmarks. Bench2Drive, Longest6 v2, and Fail2Drive are built on the CARLA simulator~\cite{dosovitskiy2017carla}, which is released under the MIT License. NAVSIM~\cite{dauner2024navsim} uses real-world driving data and is released under the Apache-2.0 License. All four benchmarks, as well as the SimLingo~\cite{renz2025simlingo} and ReCogDrive~\cite{li2025recogdrive} backbones, are publicly available for academic research. We will fully open-source our code, trained gate checkpoints, training data, and evaluation logs to support reproducibility and future research. Users who build upon our released materials should comply with the licenses of the underlying components and cite the relevant works accordingly.

\paragraph{Potential Risks}
\label{sec:appendix_potential_risk}
BLUE is evaluated through closed-loop CARLA simulation and an offline NAVSIM analysis on real-world driving data. The NAVSIM experiment does not involve testing on a physical vehicle. Deployment in real-world conditions would require additional domain-specific training and thorough safety verification to account for sensor noise, distribution shift, and safety-critical edge cases. Any real-world application should follow established engineering protocols for autonomous driving systems.

\paragraph{Broader Impact}
\label{sec:appendix_broader_impact}

This work shows that selectively generating language in VLA driving models improves both driving performance and inference efficiency. By reducing unnecessary language generation, BLUE lowers the computational cost of VLA models, making them more practical for real-world vehicle deployment where hardware budgets and energy consumption are constrained. This also reduces the carbon footprint associated with running large language models at inference time. All code, data, and evaluation logs are fully released to facilitate reproducible research on efficient language use in embodied agents. Since BLUE operates through simulation and studies when language benefits driving, we do not foresee direct negative societal consequences from this research.

\paragraph{LLM Use Statement}
\label{sec:appendix_llm_use_statement}
Large language models were used in this work exclusively for English language polishing and assisting with code implementation. All research ideas, experimental designs, analyses, and scientific conclusions are original contributions of the authors. The authors carefully reviewed all LLM-assisted outputs and take full responsibility for the final content of this paper.

\clearpage

\begin{table*}[p]
\centering
\begingroup
\footnotesize
\setlength{\tabcolsep}{4pt}
\renewcommand{\arraystretch}{1.08}
\rowcolors{2}{bbb!4}{white}
\begin{tabular}{@{}r>{\hspace{1.2em}}l>{\hspace{3em}}r>{\hspace{1.2em}}l@{}}
\toprule
\rowcolor{bbb!12}
\textbf{ID} & \textbf{Scenario} & \textbf{ID} & \textbf{Scenario} \\
\midrule
1 & MergerIntoSlowTrafficV2 & 23 & ControlLoss \\
2 & Accident & 24 & CrossingBicycleFlow \\
3 & HazardAtSideLane & 25 & HazardAtSideLaneTwoWays \\
4 & InterurbanActorFlow & 26 & VanillaSignalizedTurnEncounterGreenLight \\
5 & StaticCutIn & 27 & VanillaNonSignalizedTurnEncounterStopsign \\
6 & ParkingExit & 28 & MergerIntoSlowTraffic \\
7 & ConstructionObstacleTwoWays & 29 & T\_Junction \\
8 & OppositeVehicleTakingPriority & 30 & SignalizedJunctionLeftTurnEnterFlow \\
9 & DynamicObjectCrossing & 31 & SignalizedJunctionRightTurn \\
10 & VanillaNonSignalizedTurn & 32 & EnterActorFlow \\
11 & VanillaSignalizedTurnEncounterRedLight & 33 & NonSignalizedJunctionLeftTurn \\
12 & ParkingCutIn & 34 & ParkedObstacleTwoWays \\
13 & AccidentTwoWays & 35 & SignalizedJunctionLeftTurn \\
14 & InvadingTurn & 36 & OppositeVehicleRunningRedLight \\
15 & HighwayExit & 37 & ParkingCrossingPedestrian \\
16 & InterurbanAdvancedActorFlow & 38 & SequentialLaneChange \\
17 & NonSignalizedJunctionLeftTurnEnterFlow & 39 & VehicleTurningRoutePedestrian \\
18 & YieldToEmergencyVehicle & 40 & PedestrianCrossing \\
19 & ConstructionObstacle & 41 & BlockedIntersection \\
20 & HighwayCutIn & 42 & VehicleOpensDoorTwoWays \\
21 & NonSignalizedJunctionRightTurn & 43 & VehicleTurningRoute \\
22 & HardBreakRoute & 44 & ParkedObstacle \\
\bottomrule
\end{tabular}
\rowcolors{2}{}{}
\endgroup
\caption{Mapping between scenario IDs in Figure~\ref{fig:scenario_language_effect_all_settings} and Bench2Drive scenario names.}
\label{tab:scenario_index_mapping}
\end{table*}

\end{document}